\documentclass{article}

\PassOptionsToPackage{numbers,compress,sort}{natbib}
\usepackage[preprint]{neurips_2023}

\usepackage[utf8]{inputenc} 
\usepackage[T1]{fontenc}    

\usepackage{nicefrac}       
\usepackage{amsfonts}       
\usepackage{amsmath}
\usepackage{booktabs}           
\usepackage{multirow}           
\usepackage{graphicx}           
\usepackage{duckuments}         
\usepackage{microtype}      
\usepackage{xcolor}         
\usepackage{hyperref}
\usepackage{cleveref}
\usepackage{caption}

\usepackage[normalem]{ulem}
\useunder{\uline}{\ul}{}

\newcommand{\systemname}{BatchGNN}
\newcommand{\systemnameabbr}{BGNN}
\newcommand{\systemnameabbrc}{BGNNC}

\title{\systemname{}: Efficient CPU-Based Distributed GNN Training on Very Large Graphs}

\author{
Loc Hoang\\
KatanaGraph\\
\texttt{loc@katanagraph.com}\\
\And
Rita Brugarolas Brufau\\
Intel Corporation\\
\texttt{rita.brugarolas.brufau@intel.com}\\
\And
Ke Ding\\
Intel Corporation\\
\texttt{ke.ding@intel.com}\\
\And
Bo Wu\\
Colorado School of Mines~\thanks{Bo Wu contributed to this work while employed at KatanaGraph.}\\
\texttt{bwu@mines.edu}\\
}

\begin{document}

\maketitle

\begin{abstract}
We present \systemname{}, a distributed CPU system that showcases techniques
that can be used to efficiently train GNNs on terabyte-sized graphs.  It reduces
communication overhead with \emph{macrobatching} in which multiple minibatches'
subgraph sampling and feature fetching are batched into one communication relay
to reduce redundant feature fetches when input features are static.
\systemname{} provides integrated graph partitioning and native GNN layer
implementations to improve runtime, and it can cache aggregated input features
to further reduce sampling overhead.  \systemname{} achieves an average
$3\times$ speedup over DistDGL on three GNN models trained on OGBN graphs,
outperforms the runtimes reported by distributed GPU systems P$^3$ and
DistDGLv2, and scales to a terabyte-sized graph.
\end{abstract}

\section{Introduction}
\label{section:intro}

Graph neural networks (GNNs) apply machine learning onto graphs by generating or
transforming features on vertices and edges using deep neural networks.  Once a
GNN model is trained, it can be used in areas like fraud detection with node
classification~\cite{gnnfrauddetect}, product relationship
prediction~\cite{product_rel_predict} with link prediction, and drug discovery
using GNN-based graph generation~\cite{gnn_drug_gen}.

Training GNNs on large graphs is compute intensive.
In addition, real-world graphs like social networks or purchase graphs easily
reach terabyte scale as they can contain billions of vertices and trillions of
edges. To reduce computational and memory costs of GNN training, many GNN
systems consider a sample of neighbors in a minibatch (a batch of training
vertices instead of all vertices) to reduce the memory and computational load.
Many systems adopt this execution model because it has been observed to have a
positive effect on training~\cite{distdglv2,minibatch1} and because the smaller
memory footprint allows the use of GPUs which have limited memory compared to
CPUs.

Distributed CPU/GPU GNN systems like P$^3$~\cite{p3} and DistDGL~\cite{distdgl}
provide increased computational power and more memory in addition to
providing sampling and minibatching to efficiently train larger graphs. These
systems vary in the assumptions they make: some assume there is enough memory to
fit the entire graph on each machine while others \emph{partition} the graph
among machines. SALIENT~\cite{SALIENT} and PyTorch Geometric~\cite{PyG} fall
into the first category and can only handle small to medium-sized graphs. Systems like
DistGNN~\cite{DistGNN} and DistDGL~\cite{distdgl} are part of the second
category, and this is the approach this paper focuses on since it allows
training of arbitrarily large graphs via scale-out. One problem with this
approach is that the communication overhead during training can be
high. Unless the full computational subgraph exists locally or is fetched before
training, an updated vertex feature must be communicated to all machines that
need a copy of it. There are techniques to reduce the overhead of this
communication by allowing stale features or approximating
features~\cite{DistGNN,bns-gcn,gnnautoscale}, but they change the semantics of
training. Alternatively, some systems fetch the entire computational subgraph to
a local machine before training. This requires topology fetching and initial
feature fetching, but no communication occurs during training aside from
gradient updates.

Existing distributed GNN systems are typically bottlenecked by the communication
required to do training. For example, in P$^3$~\cite{p3}, every machine sends a
local minibatched subgraph to all other machines for computation (because the
features are divided among all machines), and the final hidden features are
reduced from all machines. This approach fails to scale: it is slower than
DistDGL if the hidden feature size of the GNN is sufficiently large~\cite{p3}
because hidden features are communicated, and more machines means more
communication partners for all processes. DistDGL, on the other hand, suffers
from excessive redundant features fetches.
In 1 epoch for a 3-layer GNN using a minibatch size 1024 on various Open Graph
Benchmark (OGB)~\cite{OGB} graphs, the number of fetched features can be more
than ten times larger than the number of vertices of the graph
(\Cref{tbl:feat_fetch} in \Cref{section:evaluation}).
Sampling bottlenecks extend to GPU GNN systems as well.  Distributed sampling
typically occurs on CPUs, so GPU systems have the same sampling overheads as CPU
systems in addition the copy cost of sampled subgraphs from CPU to GPUs.  Even
if all stages of training executed on GPUs, such a system would demand a larger
number of GPUs than CPUs since GPUs have less memory than CPUs which could lead
to a more expensive cluster than a CPU cluster.

We address these problems in \systemname{}, a distributed end-to-end GNN
training system on CPUs.  We use a sampling scheme called \emph{macrobatching}
which \emph{batches the minibatches} and communicates subgraphs and vertex
features for $M$ minibatches in one communication relay rather than $M$
relays when the features do not change during training.  This reduces redundant
feature fetching by doing it for $M$ minibatches at once.
\systemname{} accelerates end-to-end execution with an integrated distributed
streaming graph partitioner and native GNN layer implementations that avoid
dynamic memory allocation.  \systemname{} also implements an \emph{aggregation
cache} that pre-computes the aggregation of static input features which can be
fetched in lieu of doing sampling and initial feature fetching for the first GNN
layer's subgraph to reduce the overhead of sampling further. 

We compare \systemname{} against DistDGL (Section~\ref{section:evaluation})
using GraphSAGE~\cite{graphsage}, Graph Convolutional Network (GCN)~\cite{gcn},
and Graph Isomorphism Network (GIN)~\cite{GIN} models on OGBN~\cite{OGB} graphs
and show that it is $3\times$ faster on average than DistDGL on CPUs.
By replicating the experiments conducted by state-of-the-art GPU GNN systems
P$^3$ and DistDGLv2~\cite{distdglv2}, we also show that \systemname{} is able to
outperform the reported runtimes of those two GPU systems with only CPUs.

This paper makes the following major contributions:
\begin{itemize}
  \item Presents a distributed CPU-based GNN system, \systemname{}
  \item Proposes \emph{macrobatching} to lower communication overhead of
  minibatch subgraph sampling by reducing communication relays and redundant
  feature fetching
  \item Proposes memory-related optimizations to GNN layers and an input
  feature aggregation cache to elide computation for the first computational GNN
  layer
  \item Shows that \systemname{} outperforms DistDGL CPU training by $3\times$,
  outperforms reported performance of GPU systems P$^3$ and DistDGLv2, and
  trains a terabyte scale graph
\end{itemize}

\section{Background, Related Work, and Motivation}
\label{section:background}

\subsection{Graph Neural Networks}
\label{subsec:background-gnn}

Graph neural networks are an application of neural networks on the features
of vertices (or edges) of a graph.
Formally, given a graph $G$ with vertices $V$ and edges $E$, input feature
$h^{0}_{v}$ for vertex $v$, an $l$-layer GNN transforms $h^{0}_{v}$ into
$h^{l}_{v}$ through $l$ layers of \emph{aggregation} and
\emph{update}. The feature $h^{k}_{v}$ after computational
layer $k$ is defined as follows:

\begin{gather}
{a}_{v}^{k}=\operatorname{aggregate}(\{ h_{u}^{k-1} | u \in \mathcal{N}(v)\}),\ \ {h}_{v}^{k}=\operatorname{update}(h_{v}^{k-1}, a_{v}^{k})
\end{gather}

Aggregation occurs over a \emph{neighborhood} $\mathcal{N}$ of
$v$ to get the aggregated feature, and the aggregated and original features are
passed into an update function. Both steps can be parameterized. These
parameters can be \emph{trained} via supervised learning with
ground truth output for a particular input.
Many GNN architectures exist that differ in neighborhood definition, aggregation
function, and update function such as GCN~\cite{gcn},
GraphSAGE~\cite{graphsage}, GAT~\cite{gat}, and GIN~\cite{GIN}.


\subsection{GNN Systems}
\label{subsec:background-systems}

GNN systems facilitate the definition and training of GNNs.  For example,
PyTorch Geometric (PyG)~\cite{PyG} and DGL~\cite{DGL} provide graph
representations and predefined GNN layers on top of PyTorch~\cite{PyTorch}, a
neural network learning framework, to execute on CPUs and GPUs.  To reduce the
cost of training, GNN systems implement \emph{minibatching} and \emph{sampling}.
\emph{Minibatching} processes the training set in smaller minibatches instead of
all at once. This can reduce the memory footprint significantly since training
on a vertex may involve a large $l$-hop-induced subgraph for an $l$-layer GNN.
\emph{Sampling} reduces the size of the $l$-hop-induced subgraph by sampling a
fixed number of neighbors.  Different methods of sampling exist such as
GraphSAGE~\cite{graphsage}, GraphSAINT~\cite{GraphSAINT},
FastGCN~\cite{fastgcn}, layer-dependent importance sampling
(LADIES)~\cite{ladies_sampling}, and influence-based
minibatching~\cite{sampleibmb}.

Large graphs may not fit on a single machine for training.
Distributed GNN systems (the focus of this paper) allow training large graphs by
using multiple CPU and/or GPUs in parallel for additional computational power
and aggregate memory.  This facilitates training on large graphs that would be
infeasible to train on a single machine.  Existing distributed GNN systems
include Graph-Learn~\cite{AliGraph}, DistDGL~\cite{distdgl, distdglv2},
CAGNET~\cite{CAGNET}, DeepGalois~\cite{deepgalois}, SALIENT~\cite{SALIENT},
DistGNN~\cite{DistGNN}, Roc~\cite{rocgnn}, GraphScope~\cite{graphscope}, 
Euler~\cite{eulergnn}, ByteGNN~\cite{bytegnn}, NeuGraph~\cite{ma2019neugraph},
and P$^3$~\cite{p3}.  Out-of-core GNN systems also allow training large graphs
by reading data from disk on demand. They optimize disk reads for performance:
Ginex~\cite{park2022vldb}, for example, determines an optimal disk read order by
sampling multiple minibatches at once in a ``superbatch'' to determine used
features in training~\footnote{The high-level idea (sampling many batches at
once) is similar to our proposed macrobatch technique; we use it to reduce
communication.  We were not aware of superbatching when we developed our
technique.}.


\subsection{Distributed GNN Training}
\label{subsec:background-reqs}

Distributed GNN training requires graph partitioning and distributed data fetching.

\paragraph{Graph Partitioning} 

Partitioning assigns the vertices and edges of a graph uniquely among all
machines. Most distributed GNN systems use an edge-cut where 
all outgoing (or incoming) edges of a vertex get assigned to the machine that
owns that vertex. This simplifies sampling since it guarantees that
the owner of a vertex has all of its edges.
The partitioning of vertices determines the pattern of communication for
sampling and property fetching, so optimizing it is important for performance.
For instance, DistDGL~\cite{distdgl} uses METIS~\cite{METIS} to reduce the
number of edges cut (i.e., edges with an endpoint owned by another
partition) among partitions to reduce sampling/property fetching communication.
Some systems like SALIENT~\cite{SALIENT} avoid
partitioning by loading the entire graph on every machine. This allows the
system avoid communication of topology/feature data.

\paragraph{Topology Sampling and Property Fetching}

Subgraph sampling and feature fetching require communication with other
partitions. Each machine constructs a minibatch subgraph that it independently
trains, and the gradients from training are reduced among machines to update the
model.
An alternative approach is for all machines to collectively work on the same
logical subgraph rather than each sample their own subgraph (e.g.,
used by DistGNN~\cite{DistGNN}, DeepGalois~\cite{deepgalois}, and
BNS-GCN~\cite{bns-gcn}).  These systems communicate updated vertices' feature
vectors if the updated feature is not computed
locally. This does not scale because it occurs at every GNN layer unless
stale values are allowed by the system. Another approach used by P$^3$ is to
communicate the ``task'' rather than the data.  Because P$^3$ partitions all
features among all machines, each machine communicates its local subgraph to all
other machines so that the other machines apply their owned features to the
subgraphs. This also does not scale and is slower than DistDGL~\cite{p3} for
non-trivial hidden feature sizes as it requires all-to-all communication of
subgraphs and the communication of computed hidden features.  This paper focuses
on the approach in which each machine constructs an independent subgraph.


\subsection{Distributed Sampling Bottlenecks}
\label{subsec:background-motivation}

We note the following bottlenecks in distributed sampling.

\paragraph{High number of communication rounds for sampling}

Subgraph sampling requires fetching of topology from other machines with a
back-and-forth communication relay that requires many rounds of waiting on
request fulfillment.  Lowering the number of these relays is crucial for performance.

\paragraph{Redundant feature fetching}

Redundant communication of vertex feature vectors can occur over many
minibatches. For example, if a vertex A's feature is fetched in minibatch 1 and
also needs to be fetched in minibatch 2, A is redundantly fetched in minibatch
2.  The possible magnitude of redundancy is shown in \Cref{tbl:feat_fetch} where
DistDGL fetches more remote features in a single epoch than the number of
features that would exist if the graph was replicated on all machines.

\section{\systemname{} Overview}
\label{section:system}

\begin{figure*}
\centering
\includegraphics[width=0.80\linewidth]{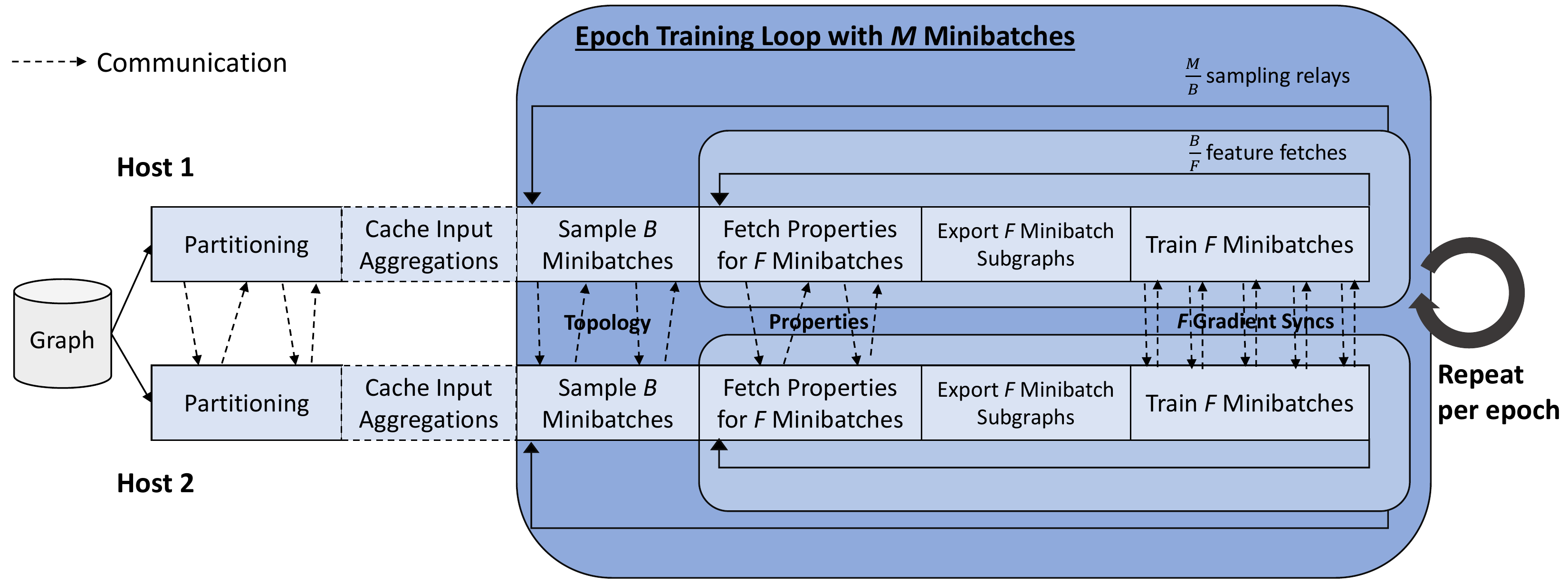}
\caption{\systemname{} execution overview. $B$ is the number of minibatches
to sample at once and $F$ is the number of minibatches to fetch properties for at
once.}
\label{fig:system_flow}
\end{figure*}

\systemname{} is an end-to-end distributed GNN system that uses
\emph{macrobatch sampling} to reduce communication overhead.
In addition, it provides integrated streaming partitioning and a subgraph exporter
that allows execution with DGL, PyG, or a set of native \systemname{} GNN layers
that are integrated into PyTorch's Autograd system. 
It also provides an input feature aggregation cache to elide the first
computational layer's sampling and feature fetching.
The core implementation is built on top of the Galois~\cite{galois} C++ parallel
library which provides graph data structures and parallel loops. Native
\systemname{} GNN layers use Intel MKL~\cite{mkl} for matrix operations.
Each machine trains on an independent sampled minibatch subgraph, and the
gradients from each minibatch are synchronized via Torch Distributed Data
Parallel.  Figure~\ref{fig:system_flow} shows the flow of \systemname{} which
is detailed below.

\paragraph{Streaming Partitioner}
\systemname{}'s partitioner builds upon the Customizable Streaming Partitioner
(CuSP)~\cite{cusp}.  As a \emph{streaming} partitioner, owner assignments of
vertices/edges are made once and not refined unlike what an offline partitioner
like METIS~\cite{METIS} does.
Partitioning occurs on-demand in a distributed fashion for however many
processes exist, and partitions do not need to be saved to disk (but it is
possible to do). Because partitioning is on-demand, the user does not need to
run an offline partitioner for every rank count configuration (e.g., DistDGL
requires offline partitioning for each number of ranks).
\systemname{} uses edge-cut partitions 
to facilitate fast topology sampling during training
(see Section~\ref{subsec:background-reqs}). \systemname{} also adds efficient feature
vector communication to the partitioning routine which did not exist in the
original CuSP system.

\paragraph{Macrobatch Sampling}

The macrobatch sampler, the key to the system's performance, reduces sampling
overhead by merging the communication that would occur over multiple minibatches
into a single ``macrobatch''. This uses the fact that if excess memory is
available on machines, it is possible to fetch the data for many minibatches at
once.  Fetched features must be static between minibatches (e.g., not learned as
part of training) for macrobatch-fetched features to remain correct; this is not
a problem for our use case where we train graphs with existing features.  An
epoch's $M$ minibatches normally requires $M$ communication relays in which the
$l$-hop sampled subgraph and its features are requested and fetched.  These
relays block the communication process to wait for request responses. This slows the
system.  Therefore, instead of $M$ relays for $M$ minibatches, \systemname{}
does one relay for $B$ minibatches at once in a \emph{macrobatch} of size $B$ to
reduce the number of communication relays by a factor of $B$.
This reduces the number of communication barriers by a factor of $B$, and
vertex properties shared by the different minibatches in the macrobatch are
fetched only once.

The semantics of training are \emph{not} changed by doing a macrobatch: all $B$
minibatched are trained the same way they would be as if you fetched each one
separately.  The topology data fetched for a macrobatch of size $B$ is the same
as if you did all $B$ batches one by one, but the features fetched are reduced
based on the overlap of features fetched among the minibatches which depends on
the vertices sampled in the macrobatch (\Cref{app:macro_perf} details a
performance model for the expected benefits).  \systemname{} also allows
fetching features in batches of $F$ (where $F \leq B$) when fetching features
for all $B$ batches at once is not possible due to memory constraints. This
results in $B / F$ feature fetching rounds instead of one. Our experiments
assume $B = F$.

After sampling, \systemname{} exports homogeneous (i.e., untyped) sampled
subgraphs into different graph representations.  \systemname{} supports three
representations: DGL~\cite{DGL}, PyTorch Geometric (PyG)~\cite{PyG}, and native
\systemname{}.  $B$ minibatch subgraphs are exported to the user's desired
format, and training proceeds for those subgraphs before the next $B$ batches
are sampled.

\paragraph{Allocation-efficient Native Layers}

\systemname{} implements native GraphSAGE~\cite{graphsage}, GCN~\cite{gcn}, 
0-$\epsilon$ GIN~\cite{GIN}, ReLU, and dropout layers that (1) reuse allocated
buffer space for tensors and (2) track information required to do the backward
phase to allow in-place operations. By reusing buffers and doing operations
in-place,  dynamic memory (de)allocation overhead is reduced.  One
limitation of buffer reuse is that the layers cannot be used with a different
input before gradient computation because it would overwrite existing
intermediate results needed for the backward phase.  In practice, GNN layers are
used only once before gradients are computed, so this is not a major problem.

\begin{figure*}
\centering
\includegraphics[width=0.75\linewidth]{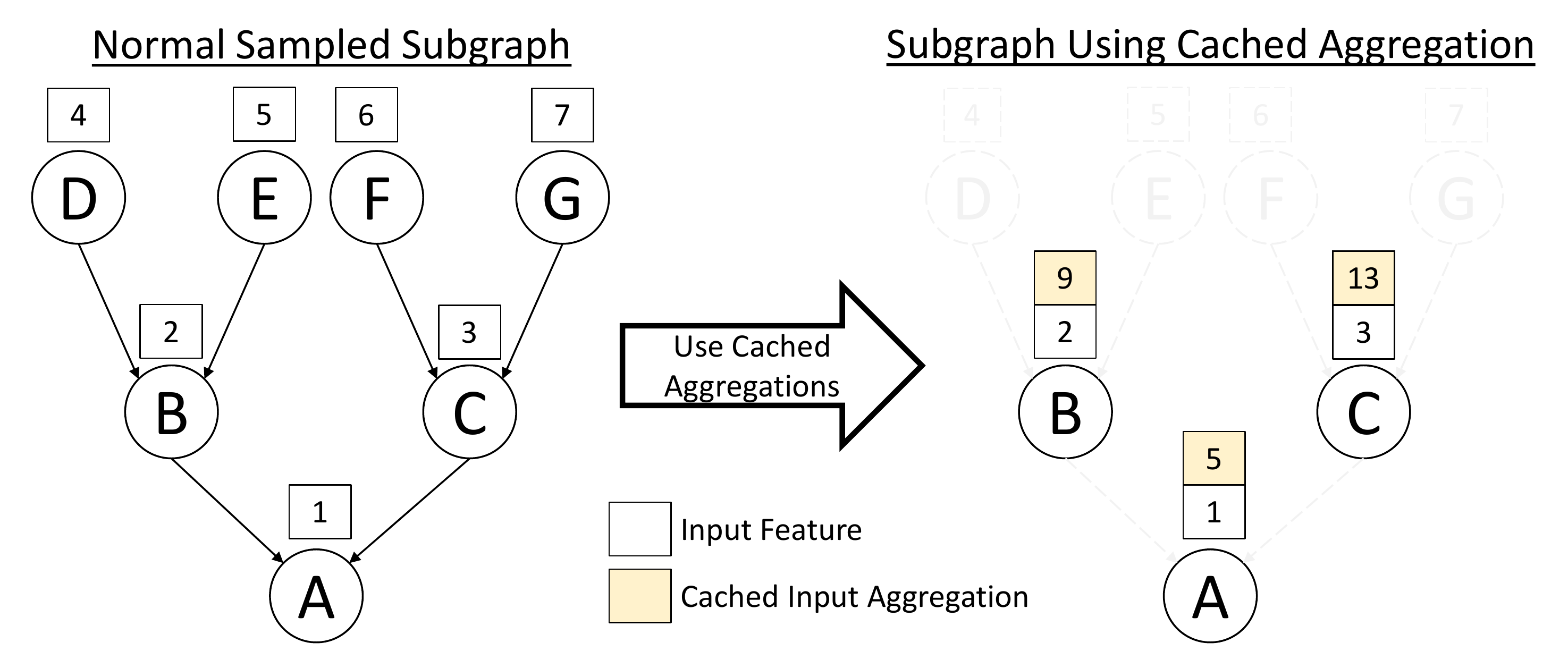}
\caption{First compute layer's subgraph with(out) cached
aggregations. Subgraph edges and the last layer's sampled nodes are not required
in a subgraph with aggregated cached values.}
\label{fig:aggregation_cache}
\end{figure*}

\paragraph{Input Feature Aggregation Cache}

\systemname{} has optional construction of an \emph{input feature aggregation
cache} that pre-computes aggregations of static input features 
for use with \systemname{} GNN layers to reduce the computation and
communication required during GNN training.  Each machine constructs a
cache which stores the aggregated feature of all neighbors
for all vertices in the graph before any sampling occurs. \systemname{}-native
GNN layers can fetch aggregated cached features of a vertex instead of doing
aggregation over neighbors during the first GNN computational layer which uses
these input features.  Because the aggregations are cached, \emph{there is no
need to construct a subgraph with edges, do sampling, or do aggregation for the
subgraph belonging to the first computational layer.}
\Cref{fig:aggregation_cache} illustrates this: the constructed subgraph with
cached aggregations only needs to fetch the aggregated feature for $A$, $B$, and
$C$ instead of the sampled nodes $D$ through $G$, and no aggregation is required
since cached features are already aggregated.  
This cache has limitations: (1) input features must be static,
(2) fetching cached features for all nodes may fetch more data than fetching
sampled node features, (3) there is initialization cost and memory overhead, and
(4) since the cache does full aggregation instead of sampled aggregation and may
estimate sampled subgraph degrees only known at runtime, training semantics are
changed.  \Cref{app:cache_discussion} and \Cref{app:cache_perf_model} contains
further discussion and a performance model for the cache.

\section{Evaluation}
\label{section:evaluation}

\subsection{Experimental Setup}
\label{subsec:setup}

The main experiments were run on up to 32 machines with a total of 72 Intel Xeon
Platinum 8360Y (Ice Lake) CPU cores spread over 2 sockets with a total of 256 GB
RAM. The machines are connected via Intel Omni-Path Architecture (OPA).

We compare \systemname{} against DistDGL (shortened to DGL henceforth)
v1.0~\cite{distdgl}; both use PyTorch 1.12.1~\cite{PyTorch}. For DistDGL, we do
not alter its default multi-threading behavior nor specify additional sampler
processes beyond the default 0. We run \systemname{} with (\systemnameabbrc{})
and without (\systemnameabbr{}) the aggregation cache and report them
separately.  The macrobatch size used by \systemname{} is large enough such that
all minibatches of an epoch are fetched in 1 macrobatch \emph{except} for papers
on 2 ranks: 2 macrobatches are used in that setting to avoid the possibility of
out-of-memory errors. Unless mentioned otherwise, the epoch time reported is for
the third epoch of execution.

We use three GNN layers: GraphSAGE~\cite{graphsage}, GCN~\cite{gcn}, and
GIN~\cite{GIN}.  For GCN and SAGE, we use 3 layers 
with a hidden feature size of 256.  For GIN, we use 3 layers
(without learnable epsilon to adjust the weight of the self-addition, i.e.,
0-$\epsilon$), and each GIN uses a 2-layer MLP with a hidden feature size of 256.
The local minibatch size for each rank is 1024, and we sample with replacement with
a sampling fan of 15, 10, 5 (where 15 is the number of samples from the initial
minibatch vertices). Evaluation of the test vertices is done with a 20,
20, 20 sampling fan (sampling for test evaluation does not heavily affect
accuracy~\cite{SALIENT}). An exception is made for GIN where we use the same fan
as the training (15, 10, 5): we observed the GIN layer is sensitive to the parameters
used during training, and changing them for testing results in poor accuracy.
During test vertex evaluation, we \emph{do not use the cache for GraphSAGE and GCN}
(meaning the fixed-neighbor aggregations that testing sees differ from
full-neighbor cached aggregations the model was trained with) but continue to
use it for GIN (the reasoning is the same as the other GIN exception before)
to keep the test comparison with DGL (which does not have a cache) the same.
Reported speedups are an average over all configurations  and GNN layers (i.e.,
including results in \Cref{app:full_results}) unless otherwise mentioned.

\begin{table}[t]
\small
\begin{minipage}{.40\textwidth}
\caption{Inputs. arxiv/papers made undirected by adding the reverse edges.}
\begin{tabular}{c|rrr}
\toprule
                             & \textbf{|V|}      & \textbf{|E|}   & \textbf{Feat.}  \\
\midrule
\textbf{arxiv}               & 0.17M             & 2.3M           & 128               \\
\textbf{products}            & 2.4M              & 123M           & 100               \\
\textbf{papers100M}          & 111M              & 3.2B           & 128               \\
\textbf{LDBC1000}            & 1.9B              & 13.5B          & 135               \\
\bottomrule
\end{tabular}
\label{tbl:inputs}
\end{minipage}
\hspace{0.025\textwidth}
\begin{minipage}{.26\textwidth}
\caption{Epoch time against P$^3$ for 16 ranks, directed papers.}
\centering
\begin{tabular}{c|r}
\toprule
\textbf{System} & \multicolumn{1}{c}{\textbf{Time (s)}} \\
\midrule
\textbf{P$^3$}  & 3.1                                   \\
\textbf{\systemnameabbr{}}    & 2.8                                   \\
\bottomrule
\end{tabular}
\label{tbl:p3}
\end{minipage}
\hspace{0.025\textwidth}
\begin{minipage}{.26\textwidth}
\caption{Epoch time against DistDGLv2 for 64 ranks, undirected papers.}
\centering
\begin{tabular}{c|r}
\toprule
\textbf{System}     & \multicolumn{1}{c}{\textbf{Time (s)}} \\
\midrule
\textbf{DistDGLv2}  & 5.0                                   \\
\textbf{\systemnameabbr{}}        & 4.5                                   \\
\bottomrule
\end{tabular}
\label{tbl:dglv2}
\end{minipage}

\end{table}

The specifications of the graphs that we experiment on are listed in
\Cref{tbl:inputs}~\cite{OGB,ldbc-snb}.  For the OGBN graphs (arxiv,
products, papers100M), the training/test splits are the defaults 
included with the OGB package~\cite{OGB}.  arxiv and papers100M (henceforth
shortened to ``papers'') are directed graphs, so we make them
undirected by adding the reverse edge for every edge.
DGL's symmetric version of
arxiv and papers remove duplicates of edges with the same source and destination
while \systemname{} keeps them (i.e., \systemname{}'s graph has more edges
than DGL).  products is already undirected. LDBC1000~\cite{ldbc-snb} is a synthetic graph with
generated features/labels used to show that \systemname{} can handle 
terabyte-scale graphs. We add random feature vectors of size 135 and make
80\% of 1\% of the vertices into training vertices. To partition the
graphs, we use a random edge-cut partition for both \systemname{} and DGL.
DGL supports METIS partitioning, but for a fair comparison with \systemname{},
we use random DGL for the main experiments.
As a special case, we show the runtime of DGL with METIS partitions (denoted
DGL-M) for papers on 32 hosts to discuss the differences with METIS.

The full experimental setup can be found \Cref{append:exp_setup}.  The Appendix
also contains more experimental results and detailed analysis to supplement the
results presented in this section.


\subsection{Epoch Time Overview}

\systemname{} achieves better performance than DGL on CPUs and GPU systems
DistDGLv2~\cite{distdglv2} and P$^3$~\cite{p3}.  Macrobatching significantly
reduces subgraph preparation time (subgraph sampling, feature fetching, and
subgraph export) and mitigates load imbalance effects on runtime.

\paragraph{Comparison with DistDGL}

\begin{table*}[tbp!]
\footnotesize
\centering{}
\caption{Average runtime (seconds) and breakdown of distributed
GNN training for 1 epoch.  Prep time includes time to prepare
a subgraph: topology sampling, feature fetching, and export.}

\label{tbl:results}
\begin{tabular}{l|l|l|rrrrrrr}
\toprule
                                    &                              &                                  & \multicolumn{7}{c}{\textbf{SAGE}}                                                                                                                                                                                                                                           \\ \cline{4-10} 
                                    &                              &                                  & \multicolumn{4}{c|}{\textbf{Subgraph Prep}}                                                                                                        & \multicolumn{2}{c|}{\textbf{Compute}}                                          &  \multicolumn{1}{c}{\textbf{Epoch}}   \\
\textbf{Graph}                      & \textbf{Ranks}               & \textbf{System}                  & \multicolumn{1}{c}{\textbf{Topo}} & \multicolumn{1}{c}{\textbf{Feat}} & \multicolumn{1}{c|}{\textbf{Export}} & \multicolumn{1}{c|}{\textbf{Total}} & \multicolumn{1}{c}{\textbf{Fwd}} & \multicolumn{1}{c|}{\textbf{Bwd}}           &  \multicolumn{1}{c}{\textbf{Total}}   \\
\midrule
\multirow{3}{*}{\textbf{arxiv}}     & \multirow{3}{*}{\textbf{4}}  & \textbf{DGL}                     & 0.76                              & 0.33                              & \multicolumn{1}{r|}{-}               & \multicolumn{1}{r|}{1.10}           & 0.25                             & \multicolumn{1}{r|}{0.48}                   &  1.87                                 \\
                                    &                              & \textbf{\systemnameabbr{}}       & 0.11                              & 0.10                              & \multicolumn{1}{r|}{0.04}            & \multicolumn{1}{r|}{0.26}           & 0.32                             & \multicolumn{1}{r|}{0.58}                   &  1.19                                 \\
                                    &                              & \textbf{\systemnameabbrc{}}      & 0.11                              & 0.07                              & \multicolumn{1}{r|}{0.03}            & \multicolumn{1}{r|}{0.22}           & 0.34                             & \multicolumn{1}{r|}{0.44}                   &  1.03                                 \\
\midrule
\multirow{3}{*}{\textbf{products}}  & \multirow{3}{*}{\textbf{8}}  & \textbf{DGL}                     & 3.21                              & 4.48                              & \multicolumn{1}{r|}{-}               & \multicolumn{1}{r|}{7.70}           & 0.84                             & \multicolumn{1}{r|}{3.30}                   &  11.88                                \\
                                    &                              & \textbf{\systemnameabbr{}}       & 0.45                              & 0.73                              & \multicolumn{1}{r|}{0.31}            & \multicolumn{1}{r|}{1.52}           & 0.71                             & \multicolumn{1}{r|}{0.86}                   &  3.12                                 \\
                                    &                              & \textbf{\systemnameabbrc{}}      & 0.65                              & 0.44                              & \multicolumn{1}{r|}{0.14}            & \multicolumn{1}{r|}{1.25}           & 0.61                             & \multicolumn{1}{r|}{0.86}                   &  2.74                                 \\
\midrule
\multirow{4}{*}{\textbf{papers}}    & \multirow{4}{*}{\textbf{32}} & \textbf{DGL}                     & 5.64                              & 21.49                             & \multicolumn{1}{r|}{-}               & \multicolumn{1}{r|}{27.13}          & 1.08                             & \multicolumn{1}{r|}{13.66}                  &  41.96                                \\
                                    &                              & \textbf{DGL-M}                   & 4.51                              & 12.55                             & \multicolumn{1}{r|}{-}               & \multicolumn{1}{r|}{17.06}          & 0.94                             & \multicolumn{1}{r|}{16.93}                  &  35.02                                \\
                                    &                              & \textbf{\systemnameabbr{}}       & 0.55                              & 2.76                              & \multicolumn{1}{r|}{0.45}            & \multicolumn{1}{r|}{3.79}           & 0.99                             & \multicolumn{1}{r|}{1.81}                   &  6.64                                 \\
                                    &                              & \textbf{\systemnameabbrc{}}      & 1.41                              & 1.06                              & \multicolumn{1}{r|}{0.15}            & \multicolumn{1}{r|}{2.65}           & 0.86                             & \multicolumn{1}{r|}{2.17}                   &  5.73                                 \\
\bottomrule
\end{tabular}
\end{table*}

\begin{table}[t]
\begin{minipage}{.52\textwidth}
\small
\caption{Number of feature vectors fetched among all machines for various OGBN graphs.}
\begin{tabular}{l|r|l|r|r}
\toprule
\textbf{Graph}                     & \textbf{Ranks}                 & \textbf{System}                           & \multicolumn{1}{c|}{\textbf{Fetched}}         & \multicolumn{1}{c}{\textbf{|V|}} \\
\midrule
\multirow{3}{*}{\textbf{arxiv}}    & \multirow{3}{*}{\textbf{4}}    & \textbf{DGL}                              & 3.9M                                          & \multirow{3}{*}{0.2M}            \\
                                   &                                & \textbf{\systemnameabbr{}}                & 0.5M                                          &                                  \\
                                   &                                & \textbf{\systemnameabbrc{}}               & 0.8M                                          &                                  \\
\midrule
\multirow{3}{*}{\textbf{products}} & \multirow{3}{*}{\textbf{8}}    & \textbf{DGL}                              & 68.0M                                         & \multirow{3}{*}{2.4M}            \\
                                   &                                & \textbf{\systemnameabbr{}}                & 12.2M                                         &                                  \\
                                   &                                & \textbf{\systemnameabbrc{}}               & 15.2M                                         &                                  \\

\midrule
\multirow{4}{*}{\textbf{papers}}   & \multirow{4}{*}{\textbf{32}}   & \textbf{DGL}                              & 463.6M                                        & \multirow{4}{*}{111.1M}          \\
                                   &                                & \textbf{DGL-M}                            & 249.2M                                        &                                  \\
                                   &                                & \textbf{\systemnameabbr{}}                & 183.6M                                        &                                  \\
                                   &                                & \textbf{\systemnameabbrc{}}               & 127.4M                                        &                                  \\
\bottomrule
\end{tabular}
\label{tbl:feat_fetch}
\end{minipage}
\hspace{0.025\textwidth}
\begin{minipage}{.44\textwidth}
\includegraphics[width=0.99\textwidth]{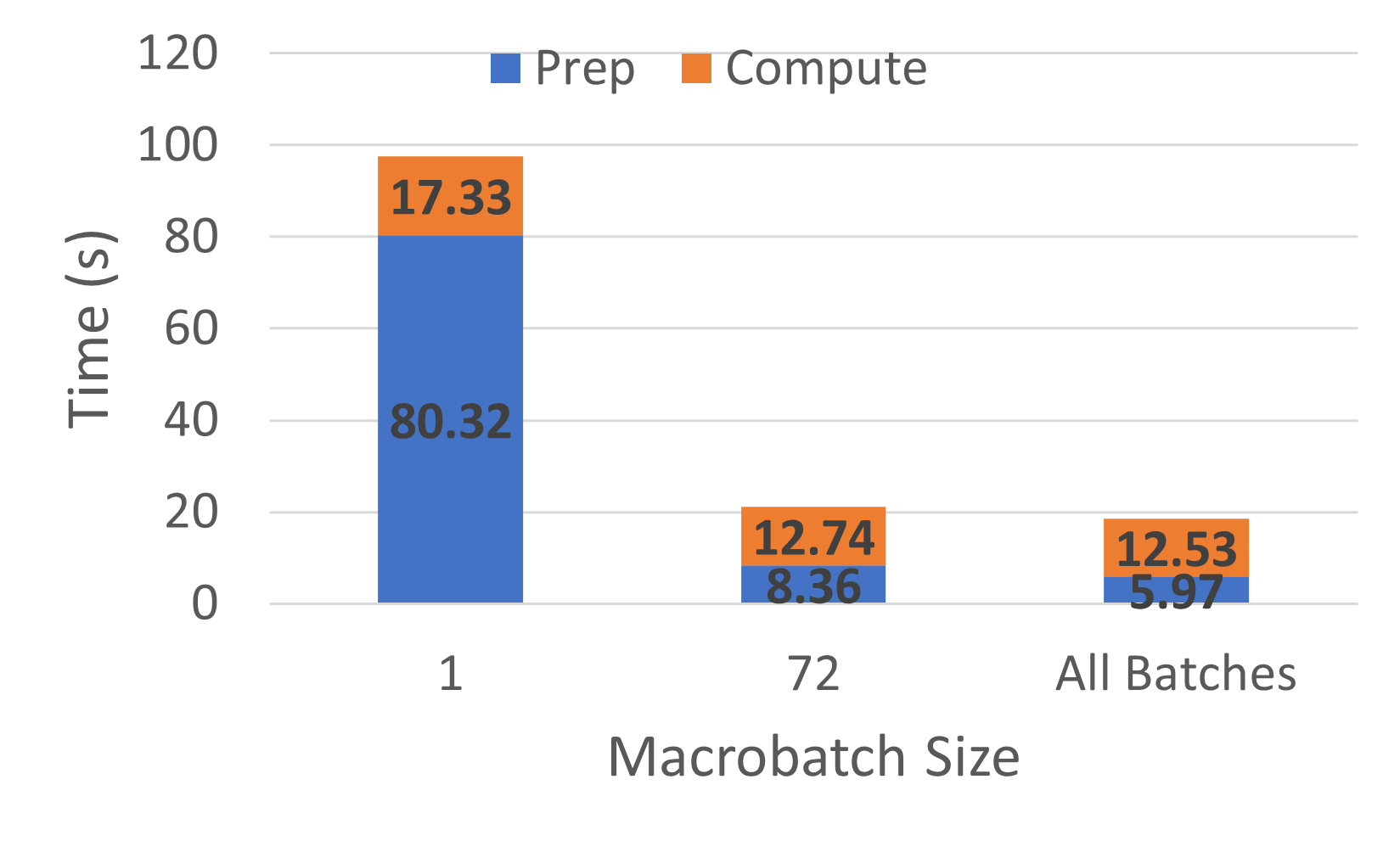}
\captionof{figure}{SAGE epoch time on LDBC1000 on 48 ranks with varying
macrobatch sizes (average among ranks of 3 runs).}
\label{fig:ldbc}
\end{minipage}

\end{table}

\Cref{tbl:results} shows the epoch time and the breakdown of training for the
OGBN graphs on DGL and \systemname{} without (\systemnameabbr{}) and with
(\systemnameabbrc{}) the aggregation cache for the SAGE layer for 4 ranks for
arxiv, 8 ranks for products, and 32 ranks for papers (\Cref{app:full_results}
contains our full results for all configurations and layers).  \systemnameabbr{}
outperforms DGL significantly: it has a geometric mean speedup of $3\times$ over
DGL over all configurations and GNN layers types.

The main reason for the speedup is macrobatch sampling.  \Cref{tbl:feat_fetch}
shows the number of features fetched in an epoch, and DGL fetches $7.8\times$,
$5.6\times$, and $2.5\times$ more features than \systemnameabbr{} does for
arxiv, products, and papers, respectively, for the settings shown in
\Cref{tbl:results}.  \systemnameabbr{} combines the  fetches of multiple
minibatches into a macrobatch to reduce redundant fetches. As a result,
\systemnameabbr{}'s fetching time is much faster than DGL's: on average,
\systemnameabbr{}'s prep time is $6\times$ faster than DGL's.

DGL-M is faster than DGL because the METIS partitioning results in smaller
sampled subgraphs and fewer remote feature fetches.  METIS places vertices to
reduce cross-partition edges which results in the seeds of a partition being
closer together in topology.  Since more neighbors are shared due to locality,
the number of sampled vertices in a subgraph decreases and results in
smaller subgraphs.  This fact combined with fewer cross-partition edges (i.e.,
subgraph consists of more local vertices) results in fewer remote
fetches (see \Cref{tbl:feat_fetch}) and lower subgraph prep time.
However, even with METIS, DGL-M is still slower than and fetches more features
than \systemnameabbr{}.

The memory optimizations that \systemname{} implements in its native GNN layers
can result in slight improvement in runtime for a larger graph like papers. For
arxiv, however, note that \systemnameabbr{}'s native layer is actually slower
than DGL, so the optimizations are not necessarily always effective.
The native layer can still be advantageous, however, because it
allows \systemname{} to avoid conversion overhead to a DGL/PyG subgraph
and integrates with the aggregation cache (\systemnameabbrc{}).

\paragraph{Comparison with GPU Systems}

\Cref{tbl:p3} and \Cref{tbl:dglv2} compare \systemname{} against runtimes
reported in the P$^3$~\cite{p3} and DistDGLv2~\cite{distdglv2} papers,
respectively. \systemname{} is run with the training setting
described in those papers to the best of our
knowledge (i.e., set the parameters we are aware of to same as
theirs).  For the P$^3$ comparison, we ran a 2-layer SAGE model with fan (25, 10),
hidden feature size 32 on 16 ranks on \emph{directed} papers with in-edge
sampling to compare against P$^3$'s run on 16 NVIDIA P100 GPUs.
For the DistDGLv2 comparison, we ran a 3-layer SAGE model with fan (15, 10, 5),
hidden feature size 256 on 64 ranks on undirected papers to compare against
DistDGLv2's run on 64 NVIDIA T4 GPUs.  For both cases, we
sample with replacement and use a minibatch size of 1000.  As shown in the
tables, \systemname{} outperforms the reported runtimes for those GPU systems.

\paragraph{Effects of Load Imbalance}

Observe in \Cref{tbl:results} that DGL's backward phase time for is
significantly higher than \systemnameabbr{}'s: the reason for this is heavy
load imbalance in execution. Imbalance in sampling runtime and forward
computation can increase backward time for all ranks if there is a single rank
slower than other ranks as backward computation involves a barrier to
synchronize gradients. We observed in our experiments that DGL has very heavy
imbalance among ranks which results in high barrier wait time. This results in
a slower epoch.  In addition, the reason that DGL-M does not result in a larger
performance improvement over DGL can be explained with imbalance as well.  We
observed METIS partitions result in more relative imbalance than random
partitions, so the backward time in \Cref{tbl:results} for DGL-M is higher than
that of normal DGL.  \systemname{} does not have this problem: its  ranks are
relatively balanced.  Any sampling imbalance only occurs when fetching
macrobatches rather than for every minibatch in the macrobatch.  Macrobatching
\emph{mitigates} the effect of imbalance because sampling does not occur past
the first minibatch of a macrobatch.  When minibatches are sampled one at a
time like in DGL, sampling imbalance reoccurs for every minibatch,
\emph{amplifying} the imbalance.  Therefore, macrobatching can improve runtime
by mitigating load imbalance overheads.  Experimental results quantifying this
imbalance and further discussion can be found in \Cref{app:imbalance}.

\paragraph{LDBC1000 and Effects of Macrobatch Size}

We run \systemname{} on LDBC1000~\cite{ldbc-snb} with added
synthetic features and labels on 48 hosts to (1) show that
\systemname{} scales out by running it on a terabyte-scale graph and (2) to
experiment with variations on macrobatch size and highlight the effects on
execution time.  In particular, we use SAGE with \systemnameabbr{}
with the same hyperparameters our main set of experiments except with variations
on macrobatch size. We select macrobatch size 1, 72, and all minibatches (e.g.
$>=$304).  \Cref{fig:ldbc} shows the results of this experiment.

Macrobatch size 1 is the slowest and has significantly higher subgraph
preparation (sampling, feature fetching, and export) time than the other two
settings.  Because sampling occurs for every minibatch, redundant features may
be fetched across minibatches since they are not fetched for multiple
minibatches at once. To illustrate this, we collected the number of remote
features vectors fetched in an epoch for these settings: macrobatch size 1
fetches 596M, macrobatch size 72 fetches 464M, and all batches fetches 363M.
Macrobatching reduces the total number of features that need to
be fetched.  Additionally, imbalance in subgraph preparation is amplified at
lower macrobatch sizes as discussed previously. This may result in slower
compute time due to imbalance and the barrier that exists for gradient
synchronization.  Most importantly, \systemname{} cannot fully leverage
parallelism during subgraph preparation in the single minibatch setting because
parallelism is interbatch: if there is only a single minibatch to prepare, there
will be single threaded execution during subgraph preparation (\Cref{app:inter_vs_intra}
discusses this implementation detail).  At macrobatch
size 72, there are 72 minibatches for the 72 threads on the machine to work on
at the same time. Combined with less redundant feature fetching, macrobatch size
72 has significantly improved runtime over the size 1 setting.  Finally, the
all-batch setting reduces the communication overhead further and results in the
fastest run in this experiment. 

\paragraph{Runtime with Aggregation Cache}

\systemnameabbrc{} is \systemname{} with the aggregation cache enabled, and as
shown \Cref{tbl:results}, it typically reduces the runtime of the epoch by
reducing the time taken by subgraph preparation. On average, an epoch is
$1.2\times$ faster in \systemnameabbrc{} than \systemnameabbr{}. Instead of
sampling and fetching for the last sampling layer, the system fetches remote
aggregated features. This remote fetch time is included in the ``Topo'' column
of the preparation time, and the change in the number of input features fetched
is reflected in the reduction of the time in the ``Feat'' column for
\systemnameabbrc{}.

The cache is not guaranteed to significantly reduce epoch time.
\Cref{tbl:feat_fetch} explains this: for arxiv and products, the cache variant
fetches more features than the no-cache variant. This is possible if sampling
does not add many new vertices since the aggregation cache must fetch aggregated
features for all vertices instead of fetching only non-cached features from
newly sampled nodes (see \Cref{app:cache_perf_model} for the performance model).
Fetching more features means (1) more communication overhead during the subgraph
preparation and (2) potentially more feature reads during the forward phase when
fetching the aggregated cached values than reads that would have occurred in the
non-cached case. Both can add enough overhead such that it outweighs the
benefits of skipping the aggregation step and reducing export time.

\begin{figure*}[tb!]
\centering
\begin{minipage}{0.32\linewidth}
\centering
\includegraphics[width=0.95\linewidth]{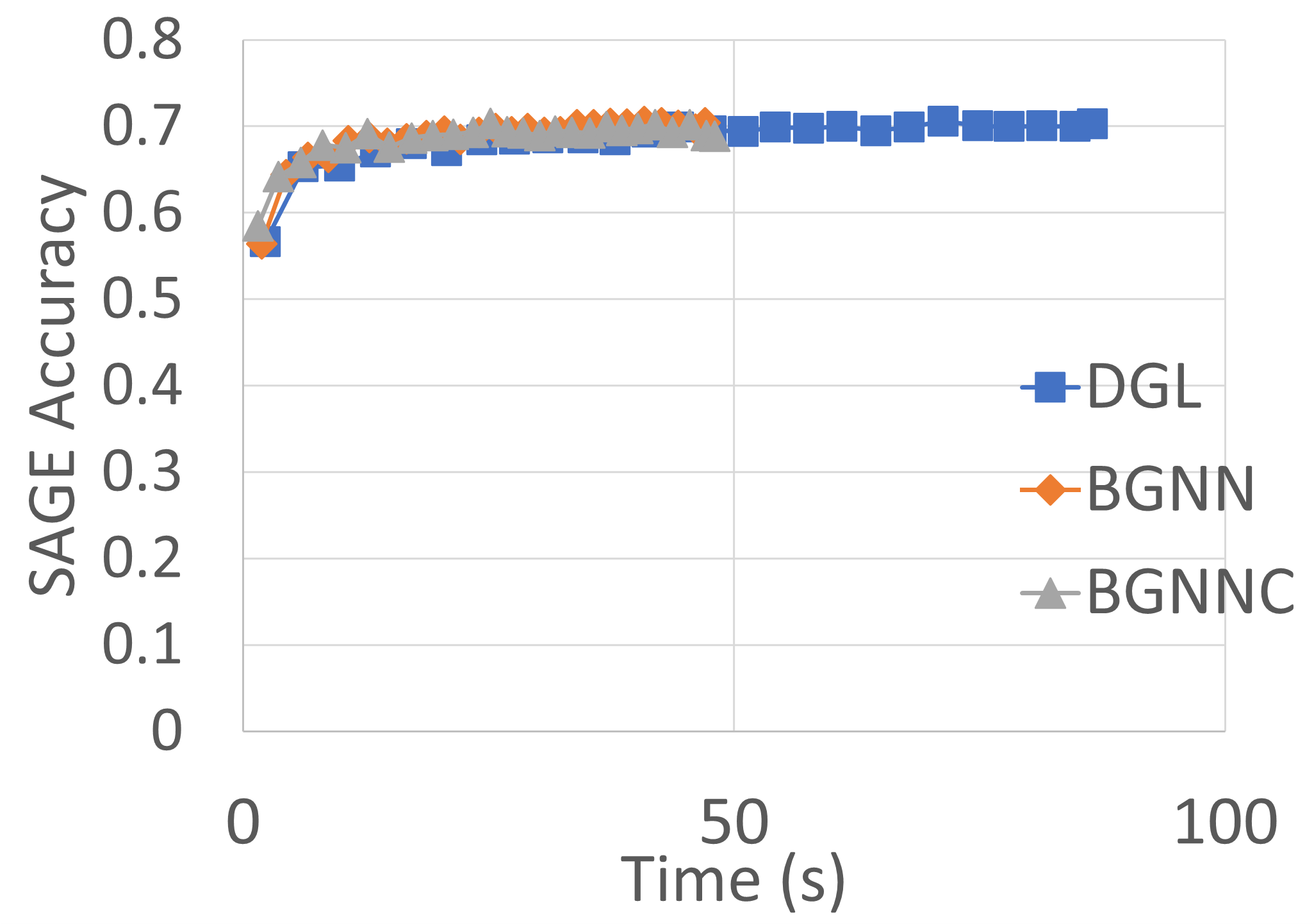}
\end{minipage}
\begin{minipage}{0.32\linewidth}
\centering
\includegraphics[width=0.95\linewidth]{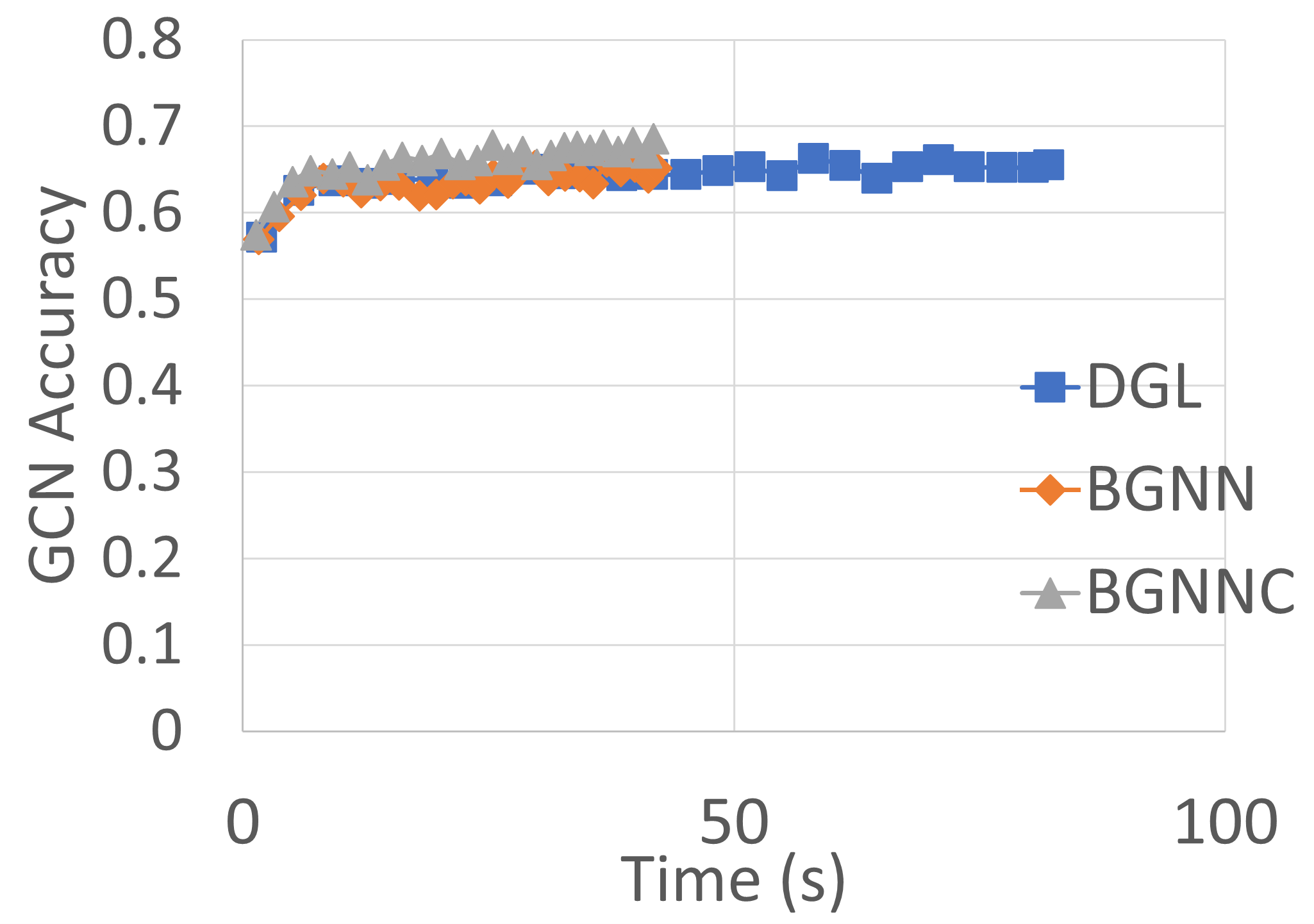}
\end{minipage}
\begin{minipage}{0.32\linewidth}
\centering
\includegraphics[width=0.95\linewidth]{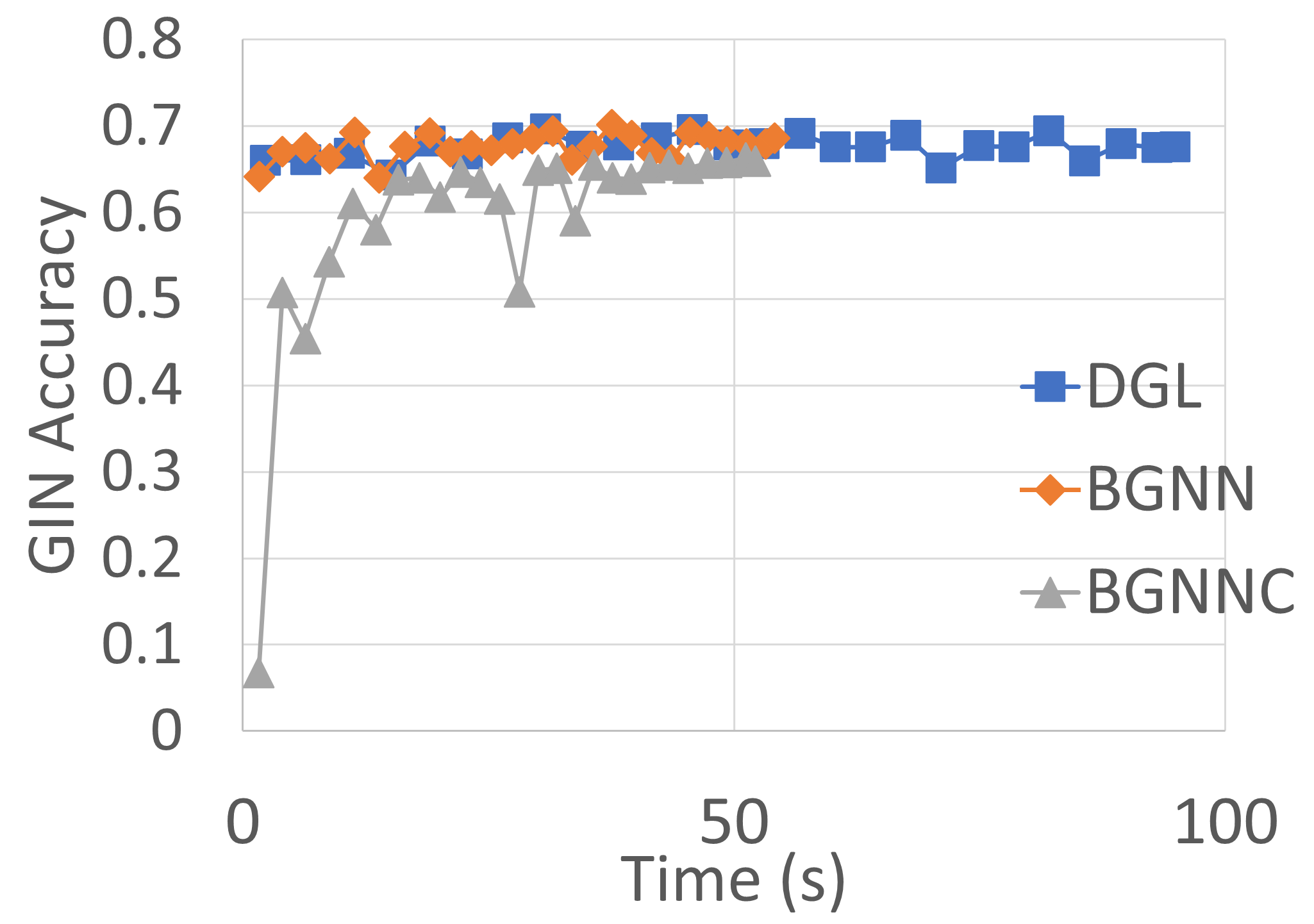}
\end{minipage}
\caption{Time (sec) to accuracy, arxiv, 4 ranks, for SAGE, GCN, and GIN, respectively.}
\label{fig:acc_arxiv}
\end{figure*}

\begin{figure*}[tb!]
\centering
\begin{minipage}{0.32\linewidth}
\centering
\includegraphics[width=0.95\linewidth]{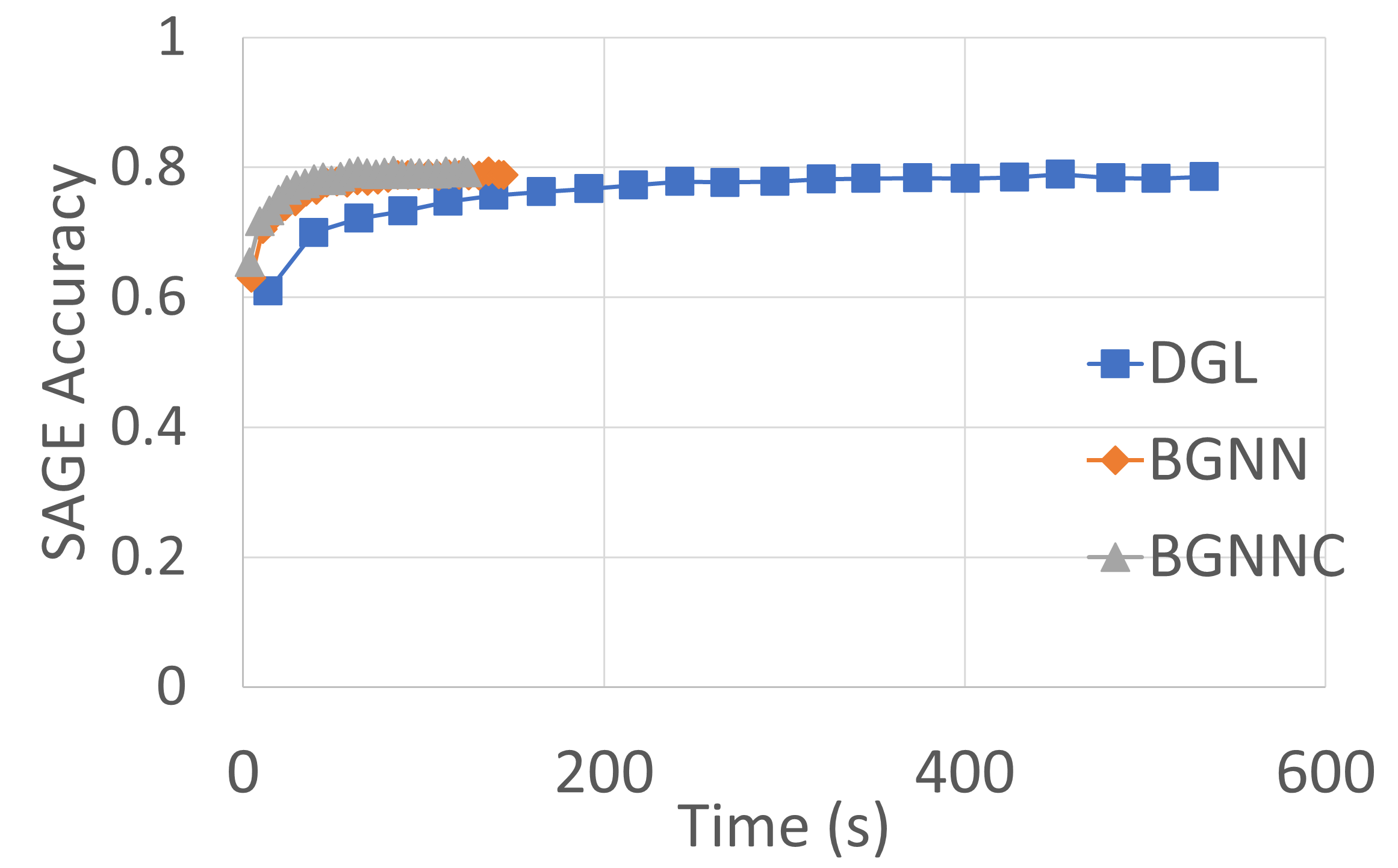}
\end{minipage}
\begin{minipage}{0.32\linewidth}
\centering
\includegraphics[width=0.95\linewidth]{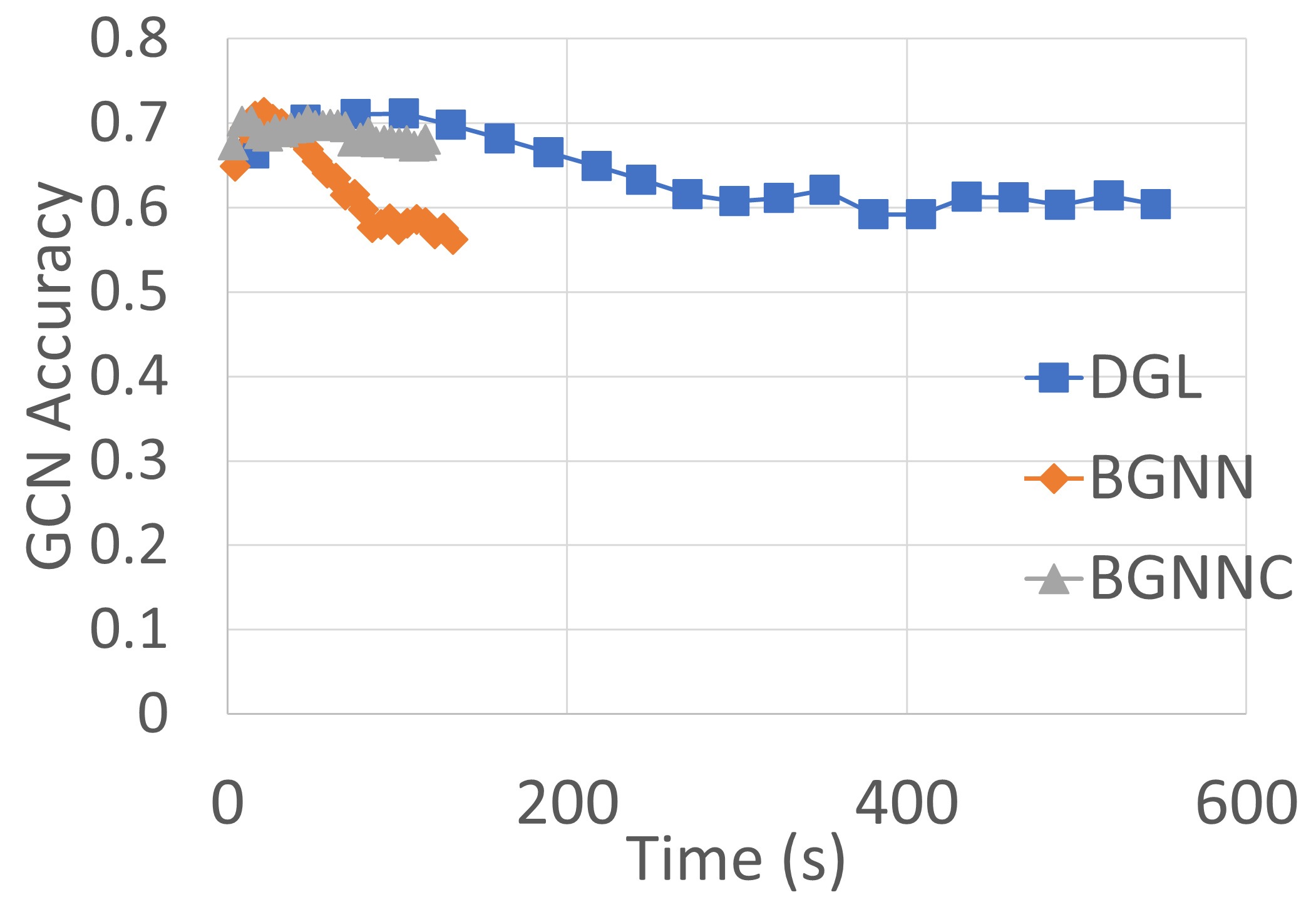}
\end{minipage}
\begin{minipage}{0.32\linewidth}
\centering
\includegraphics[width=0.95\linewidth]{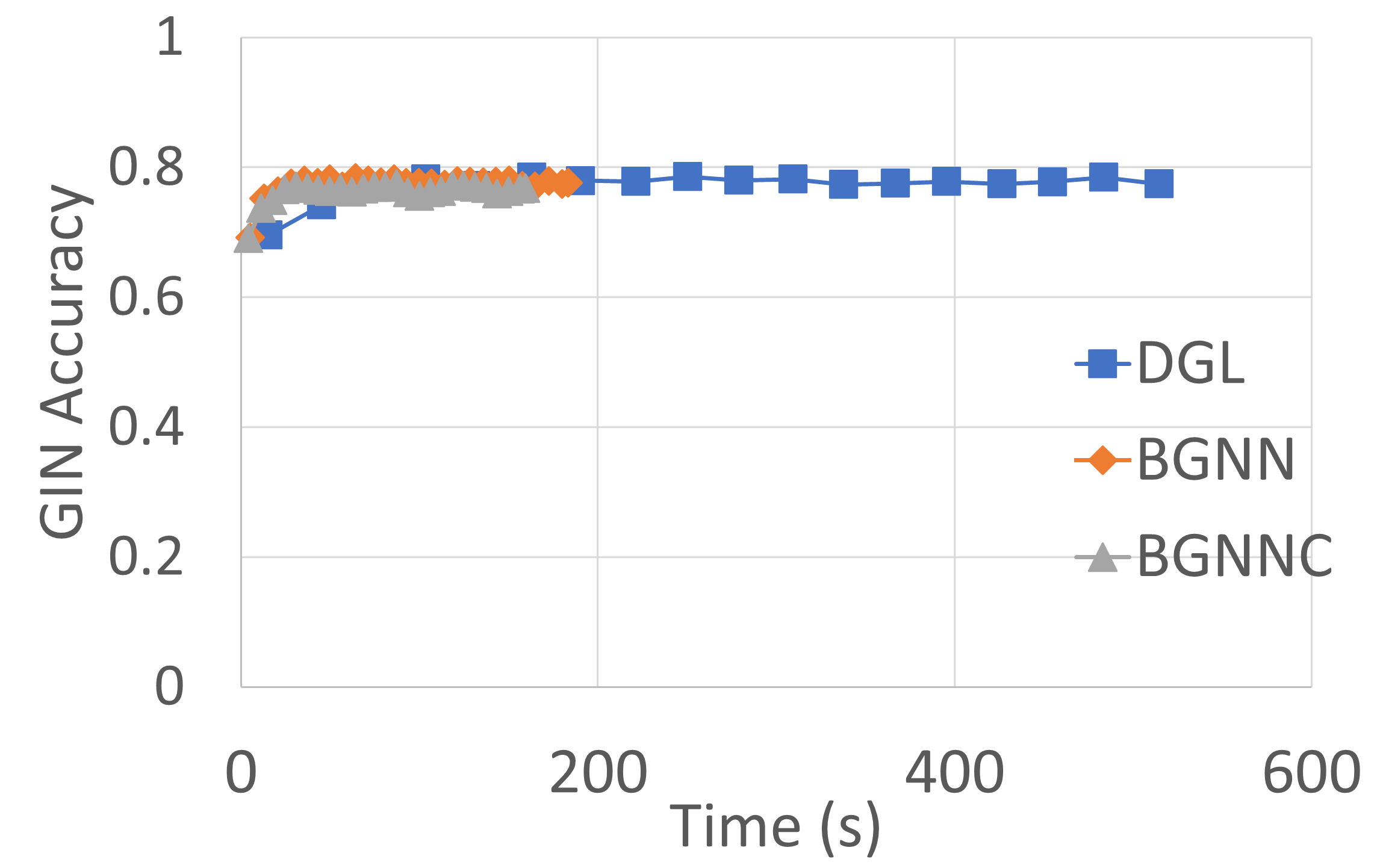}
\end{minipage}
\caption{Time (sec) to accuracy, products, 8 ranks, for SAGE, GCN, and GIN, respectively.}
\label{fig:acc_products}
\end{figure*}

\begin{figure*}[tb!]
\centering
\begin{minipage}{0.32\linewidth}
\centering
\includegraphics[width=0.95\linewidth]{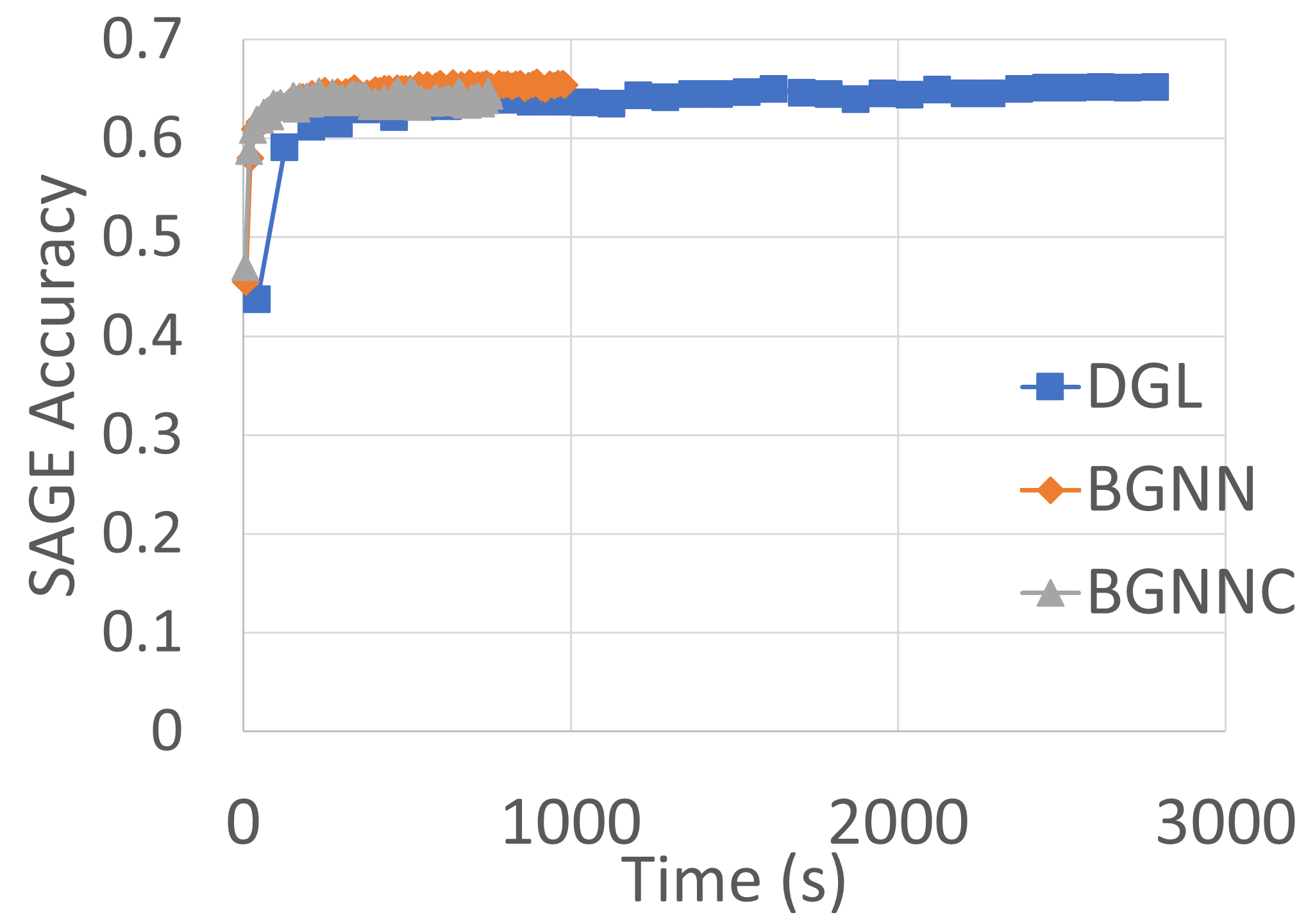}
\end{minipage}
\begin{minipage}{0.32\linewidth}
\centering
\includegraphics[width=0.95\linewidth]{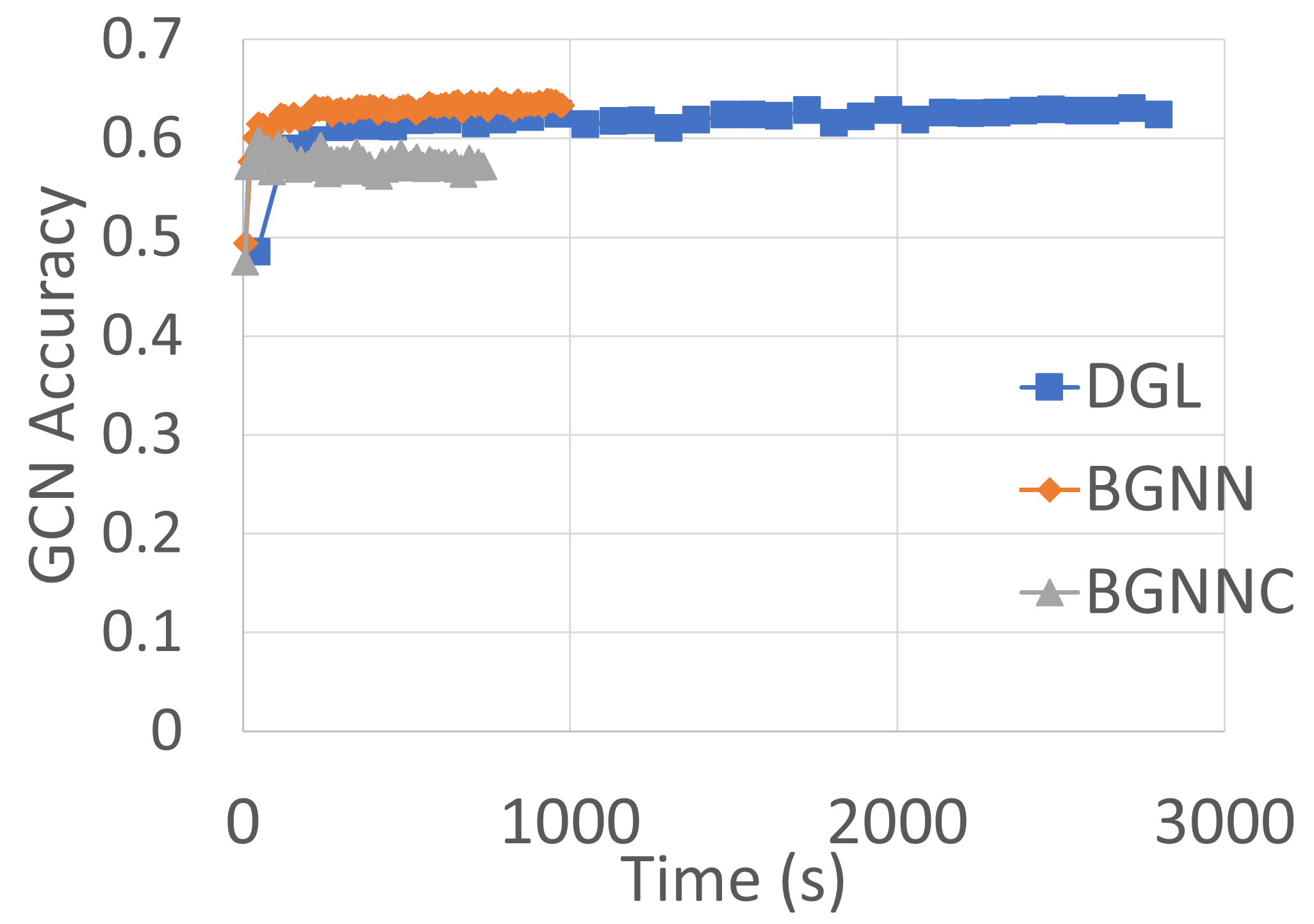}
\end{minipage}
\begin{minipage}{0.32\linewidth}
\centering
\includegraphics[width=0.95\linewidth]{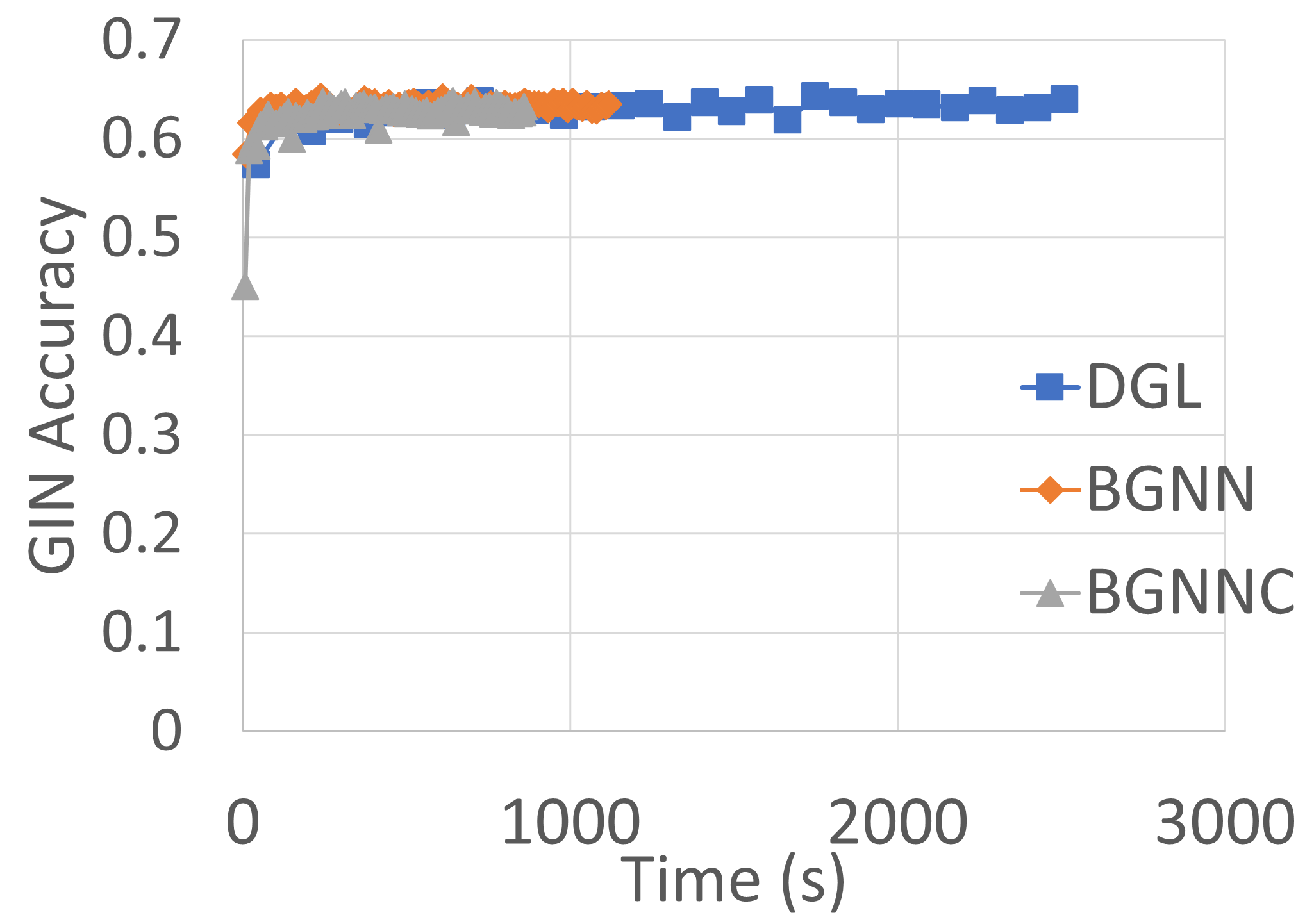}
\end{minipage}
\caption{Time (sec) to accuracy, papers, 32 ranks, for SAGE, GCN, and GIN, respectively.}
\label{fig:acc_papers}
\end{figure*}

\subsection{Time to Accuracy}
\label{subsec:tta}

\Cref{fig:acc_arxiv}, \Cref{fig:acc_products}, and \Cref{fig:acc_papers} show
accuracy plots for each graph and model (we drop some later points for DGL
because it takes longer to run the same number of epochs).  We did not optimize
hyperparameters for accuracy: these results are to compare convergence behavior
of the systems.  \systemnameabbr{} achieves similar accuracy to DGL at a faster
rate.  Minor convergence differences are expected due to the randomness of
sampling and training as well as possible low-level implementation differences
in the GNN layers (high-level semantics are the same).

Since \systemnameabbrc{} uses full-neighbor aggregation and estimated
normalization for GCN, the convergence of \systemnameabbrc{} differs from
that of \systemnameabbr{}. The effect on accuracy depends on setting it is run under:
there are instances of it improving accuracy, maintaining accuracy, and
decreasing accuracy. For example, the cache increases accuracy for GCN on arxiv,
but it decreases it for GIN on all graphs.  Therefore, the aggregation cache's
effectiveness should be evaluated on a case-by-case basis: in exchange for a
change in semantics and accuracy change, the runtime of an epoch can be
decreased.

\section{Conclusion}
\label{section:conclusion}

We presented \systemname{}, an scalable end-to-end distributed GNN training
system for CPUs that achieves state-of-the-art performance. Macrobatching does
sampling and feature fetching of many minibatches in a single communication
relay to reduce the number of communication relays and redundant feature
fetches.  In addition to macrobatching, it implements an aggregation cache to
reduce sampling costs for the first computational layer, integrated on-demand
graph partitioning, and memory-aware native GNN layers which all serve to reduce
end-to-end training time.  It is $3\times$ faster than DistDGL, outperforms
distributed GPU GNN systems P$^3$ and DistDGLv2 using only CPUs, and permits
terabyte-size graph training via scale-out.  The ideas presented in
\systemname{} can be applied to existing distributed CPU and GPU GNN systems
that use sampling to accelerate training.

\bibliographystyle{acm}
\bibliography{reference.bib}


\newpage

\appendix

\renewcommand{\thefigure}{A\arabic{figure}}
\renewcommand{\thetable}{A\arabic{table}}
\setcounter{table}{0}
\setcounter{figure}{0}
\setcounter{page}{1}

What follows is the Appendix for the paper \systemname{}: Efficient
CPU-Based Distributed GNN Training on Very Large Graphs.

\section{Macrobatching Performance Model}
\label{app:macro_perf}

This section analyzes the potential performance gain of macrobatching.

Let $C$ be the incidental data cost (e.g., confirmation messages, metadata,
etc.) of doing a communication relay for subgraph sampling. Let $T_i$ be the
size of the topology data for minibatch $i$. Let $f_i$ be the set of vertices to
fetch for minibatch $i$.  Assume for the sake of worst-case analysis that all
node features are fetched from another machine. Let S be the size of the
initial feature vector.  For a total of $M$ minibatches, the communication cost
is the following:

\begin{equation}
\sum_{i=0}^{M-1}{T_i} + MC +\sum_{i=0}^{M-1}{S * count(f_i)}
\end{equation}

Let $B$ by the macrobatch size.  Let $A_k$ be the union of all vertices to fetch
for minibatches $k*B$ through $(k+1)*B -1$ (i.e., union of $B$ minibatches).

\begin{equation}
\sum_{i=0}^{M-1}{T_i} +
\frac{MC}{B} +
\sum_{k=0}^{\frac{M}{B} - 1}{S * count(A_{k})}
\end{equation}

The incidental data cost is reduced by a factor of $B$, and the feature
data is reduced by the amount of overlap in features of the $B$ minibatches
in the macrobatch. Therefore, if there is a high overlap,
macrobatching significantly reduces sampling overhead. 
The reduction in the number of relays (i.e., barriers among
all machines) and incidental data costs provide benefit as well.

\section{Further Aggregation Cache Discussion}
\label{app:cache_discussion}

The cache is constructed once and can be reused indefinitely if the input
features to a GNN model never change and if the graph does not add new nodes
and/or edges. Otherwise, cache elements may become invalid and would need to be
updated.  \systemname{} currently only trains static graphs without learnable
inputs, so these problems are avoided.  The aggregation cache can only be used
with \systemname{} native layers since the layer must be aware that the cache
exists to use it. Other GNN layers from other systems like DGL would need to be
modified to allow the use of the cache.

In addition, training semantics change with the aggregation cache. 
Instead of aggregating over a sampled set of neighbors during compute, the
cached aggregated value which results from aggregating over \emph{all}
neighbors is used instead. As a result, training changes since the aggregated
value for the first GNN layer will always be the same instead of being dependent
on the sampled neighbors (i.e., no variation during training). 
The result is a semantically different hybrid computational scheme where the first
computational layer uses full neighbor aggregation (theoretically, more
neighborhood information is used) while the following layers use sampled
aggregation.  Also, some types of aggregation normalization like symmetric
normalization used by GCN require knowledge of the subgraph at \emph{runtime}
which is not known at cache construction time. Specifically, the degrees
of nodes being aggregated from are not known until the graph is sampled,
and this metric is required for correct symmetric normalization.
In our implementation, we estimate this value at cache construction
time using global degrees. This will change training semantics.

\section{Aggregation Cache Performance Model}
\label{app:cache_perf_model}

We analyze the possible performance gain with the aggregation cache.

Let $V$ be the set of vertices that need to be sampled from to create the
subgraph for the first computational layer, $S$ be the size of the initial
feature vector, $\mathcal{N}(V)$ be the set of vertices not in $V$ that are
selected as a result of neighbor sampling for the layer, $T_v$ be the size of
the vertex data for vertices in $V$, $T_n$ the size of the vertex data for
vertices in $\mathcal{N}(V)$, and $T_e$ be the size of the edge data for the subgraph.

The data cost for constructing the subgraph includes the input features
for all vertices in the graph and the topology data:

\begin{equation}
  VS +  count(\mathcal{N}(V)) * S  + T_v + T_n + T_e
\end{equation}

When using the cache, the data cost is instead the following.

\begin{equation}
  VS +  VS + T_v
\end{equation}

The initial input features of $V$ are still be fetched because some layers
require it, but the cost of fetching sampled vertices, their features, and
sampled edges is eliminated since they do not need to be sampled nor created. 
Instead, all cached aggregated features are fetched (the $VS$ term).  Therefore,
the aggregation cache reduces communicated data volume if $V$ is smaller than
$\mathcal{N}(V)$.
There are scenarios where this is not the case: for example, if the number of
newly sampled nodes is low because all nodes have already been sampled by the
last layer, then using the cache requires communication of significantly more
data.  Regardless of the data savings, the use of the cache eliminates the
sampling phase as well as the aggregation step in the first GNN layer. This can
have beneficial performance implications. 

\section{System Implementation Details}
\label{app:impldetails}

This section goes into detail of selected aspects of the implementation of
\systemname{}'s macrobatch sampler to allow for better understanding of the
system's performance.

\subsection{Parallel Topology Sampling}
\label{subsec:topo_sample}

Efficient neighborhood sampling is accomplished by preallocating as many
communication buffers and edge buffers as possible to allow multiple
threads to write their individual sampling results without the need for dynamic
memory allocation.  The high-level steps involved in \systemname{}'s
neighborhood sampling for a layer are (1) determining vertices to sample (known
as \emph{seed vertices}) and the machines that own the edges for them (2) making
requests to those machines to sample those vertices, (3) fulfilling the
sampling requests and sending the sampled neighbors and edges back to the
requester, and (4) processing the fulfilled requests. This subsection will
highlight selected areas of this pipeline.

\subsubsection{Generating Request Messages}
\label{subsubsec:requestmessage}

Given a set of unique vertices for each of the $B$ minibatches in a macrobatch,
\systemname{} determines which machine owns the edges for each vertex and
constructs a request message for each machine in parallel via an
inspector-executor pattern.  A Galois~\cite{galois} parallel loop construct is
used to loop over each minibatch's seed vertices in parallel. Each thread notes
the owner machine of the vertex it inspects, and a thread-local count for the
number of vertices to request from that machine is incremented. Once this first
loop concludes, memory for a message buffer for each machine is allocated as the
system knows exactly how many requests to make to each machine by summing the
thread-local counts. The minibatches are then iterated over again in parallel
using the same thread assignment as the inspector phase, and by using
the thread-local counts, each thread can serialize vertex IDs into disjoint
locations of the message buffer for a machine because the exact number each thread
writes for each machine is known from inspection. 
The minibatch ID and the seed index for the vertex being serialized for index
$i$ of the buffer is maintained while doing this serialization, to associate
each vertex with the location to store its results later.

If the input feature aggregation cache is enabled for use with native
\systemname{} GNN layers, sampling does not occur for the last sampling layer
(i.e., the first computational layer).  Instead, the system fetches the required
aggregated features for that layer's seed vertices, eliminating the need for
sampling and compute during computation (see \Cref{section:system} of the
main paper).

\subsubsection{Sampling Request Fulfillment}

Machines receive sampling requests that can be handled
efficiently by preallocating all necessary memory for the response 
in parallel. Upon receipt of a request message,
the memory for the response message is allocated based on the number of vertices
requested. The memory is a known amount because (1) the sampling is
fixed-neighbor, so the (maximum) number of neighbors is known or (2) all
neighbors are being requested, and the degree of the vertex is known. Each
requested vertex ID is then added to a thread-safe Galois sampling worklist which 
consists of requested vertices from all machines along with a pointer into the response
message memory designating where sampled vertices and edges are to be written.
This ensures that there is no redundant copying since results are written
directly into the buffer to be sent and that the thread working on that vertex
will not interfere with the writes from other threads.
Once the response buffers are allocated and the worklist is
constructed, parallel topology sampling occurs over the worklist. Each thread
takes a worklist item, does sampling for the item's vertex, and writes the
results into the item's saved pointer. The result is
parallel sampling for all minibatches in the macrobatch among all machines in
the network.  Graph access locality can also be better leveraged 
since sampling occurs in one macrobatch rather than across $B$
individual minibatches.  Once finished, each machine sends back the sampled
vertices and edges in the same order that the requester requested the vertices.

\subsubsection{Processing Received Sampled Vertices and Edges}

Because the minibatch and seed indices were saved when generating the request
message (\Cref{subsubsec:requestmessage}), it is simple to deserialize the
sampled results from other machines into the correct locations upon receipt
since are sent back in the same order. A parallel loop is used to iterate over
each request response, and the index of the vertex being processed in the buffer
links it to the minibatch and seed index to write the sampled results into based
on the agreed-upon order of messages.  The storage for result deserialization
will have been preallocated if the number of sampled neighbors is fixed. When
sampling does not involve a fixed neighbor such as in full-neighbor sampling
where all neighbors of a vertex are fetched, this storage is allocated when the
degrees of the vertices being sampled are sent to the requester (the degrees are
not sent if they are not required).

\subsection{Property Fetching and Subgraph Export}
\label{subsec:fetch_export}

Properties (e.g., labels, features) are fetched in batches on demand whenever an
exported subgraph for a minibatch is requested. When a single minibatch subgraph
is requested, a batch is created at once. This subsection describes how both
occur.

\subsubsection{Property Fetching}
Given a macrobatch with $B$ minibatches, \systemname{} fetches properties for
$F$ of those minibatches in a batch to avoid fetching redundant properties among
the subgraphs of the minibatches and to reduce the number of communication
barriers required.
First, a set of all unique vertices (and edges, if required) is compiled
in parallel by threads checking all minibatches and adding their vertices to a
concurrent set data structure. This set is used to do distributed feature
fetching: an all-to-all fetching operation pulls all remote properties
as well as local properties for the vertices in the set into a single array for
each fetched property. A mapping from vertex ID to its corresponding index in
these arrays is created as well. At this point, the properties for all entities
in the $B$ minibatched subgraphs exist on each machine.

\subsubsection{Subgraph Export}
\systemname{} exports $F$ subgraphs
(the number of minibatches that features are fetched for)
in parallel to a structure in which it can be used with the \systemname{}
native layer implementations or to a structure that is more friendly
for further transformation into a DGL or PyTorch Geometric (PyG) subgraph 
(we will refer to it as a Python subgraph) for
use with those systems.
For a native layer subgraph, the sampled edges are exported into a compressed
sparse row (CSR) format for efficient parallel iteration over individual
vertices' edges, and the transpose (a compressed sparse column, CSC) graph is
created as well to do reverse aggregation in the backward phase.  For
the Python subgraph, the edges are placed into an Arrow tensor objects 
because Arrow tensor objects can be zero-copied to PyTorch tensors which 
DGL/PyG use for edges. The Python subgraphs undergo further processing as
necessary to convert them to the DGL/PyG graph format upon use.
All subgraphs also require that the initial vertex features be placed into a
contiguous array as well. The features for the subgraph's vertices are copied
from the $F$-minibatch property array using the mapping from ID to index to
select the correct features. The features are stored in an Arrow array for
zero-copy to PyTorch.

\subsection{Discussion: Inter- vs Intrabatch Parallelism}
\label{app:inter_vs_intra}

\systemname{}'s sampler uses interbatch parallelism (threads work on
individual minibatches) rather than intrabatch parallelism (threads 
work on the same minibatch). If there are a low number of
minibatches in a macrobatch, all threads may not be fully utilized.
\systemname{} is not expected to perform well in this scenario.

Minibatches are typically small, so there is not much benefit assigning many
threads to work on a single minibatch.  In addition, intrabatch parallelism may
require the use of atomics, locks, or additional thread-safe data structures 
to allow multiple threads to work on a task at once in a minibatch. For
example, in construction of the CSR for native layer export, the counting of the
number of edges for each vertex would need to be an atomic variable which would
likely outweigh the benefits of intrabatch parallelism.  By allowing threads to
work on disjoint subgraphs of a macrobatch, more work can be 
accomplished faster with better thread utilization, and additional overhead
from thread-safe constructs is avoided. 

\section{Full Experimental Setup}
\label{append:exp_setup}

The following is the unabriged experimental setup with none of the details
omitted for purposes of space limits. Portions of this text can also be found
in the main paper.

The main experiments were run on up to 32 machines with a total of 72 Intel Xeon
Platinum 8360Y (Ice Lake) CPU cores spread over 2 sockets with a total of 256 GB
RAM. The machines are connected via Intel Omni-Path Architecture (OPA).

We compare \systemname{} against DistDGL (shorted to DGL henceforth) v1.0~\cite{distdgl}. Both use
PyTorch 1.12.1~\cite{PyTorch}. Every machine runs a single trainer process
(i.e., 1 rank per machine) for both systems.  In DistDGL's case, we do not alter
its default multi-threading behavior, and we do not specify additional sampler
processes beyond the default setting of 0. We run \systemname{} with (\systemnameabbrc{})
and without (\systemnameabbr{})
the aggregation cache and report them separately. The macrobatch size used by
\systemname{} is large enough such that all minibatches of an epoch are fetched
in 1 macrobatch \emph{except} for papers on 2 ranks: 2 macrobatches 
are used in that setting to avoid the possibility of out-of-memory errors.
We do not show results for \systemname{}'s export into DGL/PyG subgraphs
with computation with their respective layers because the key
contributions of our system are the sampler and the native layers. 
Reported numbers are not the first epoch of execution but the third because the
first epoch can be slower than the rest. For \systemname{} in particular, the
first epoch is when the initial reallocation of memory occurs for the native
layer.  On average, the third epoch for \systemname{} is $1.27\times$ faster
than the first epoch for the main experiments of this paper, and at worst, the
third epoch is $1.87\times$ faster than the first.  Unless otherwise mentioned, times
are average across all ranks of an epoch for each category (prep, forward,
etc.).

We use three GNN layers: GraphSAGE~\cite{graphsage}, GCN~\cite{gcn}, and
GIN~\cite{GIN}.  For GCN and SAGE, we use 3 layers (with ReLU/Dropout for the
first two layers) with a hidden feature size of 256.  For GIN, we use 3 layers
(without learnable epsilon to adjust the weight of the self-addition, i.e.,
0-$\epsilon$), and each GIN uses a 2-layer MLP with a hidden feature size of 256
with batch normalization and ReLU after the first layer of the MLP. Each GIN layer does another batch
normalization followed by a ReLU at the end.  We do not optimize the hyperparameters for
accuracy because evaluating the accuracy of different models is not the purpose
of this work.  However, we do report accuracy to make sure that the native layer
implementations in \systemname{} result in similar accuracy to DistDGL.  The
local minibatch size for each rank is 1024, and we sample with replacement with
a sampling fan of 15, 10, 5 (where 15 is the number of samples from the initial
vertices of the minibatch).  Evaluation of the test vertices is done with a 20,
20, 20 sampling fan (sampling for test evaluation does not heavily affect
accuracy~\cite{SALIENT}). An exception is made for GIN where we use the same fan
as the training (15, 10, 5): we observed the GIN layer is sensitive to the parameters
used during training, and changing them for testing results in poor accuracy.  Using
the aggregation cache means that the last sampling layer's fan (5) is
ignored in favor of the use of the full-neighbor cached aggregated value during
training. During test vertex evaluation, we \emph{do not use the cache for GraphSAGE and GCN}
(meaning the fixed-neighbor aggregations that testing sees differ from
full-neighbor cached aggregations the model was trained with) but continue to
use it for GIN (the reasoning is the same as the other GIN exception before)
to keep the test comparison with DGL (which does not have a
cache) the same when it does not heavily affect accuracy like GIN.

The specifications of the graphs that we experiment on are listed in
\Cref{tbl:inputs}~\cite{OGB,ldbc-snb} of the main paper.  For the OGBN graphs (arxiv,
products, papers100M), the training/test splits are the defaults 
included with the OGB package~\cite{OGB}.  arxiv and papers100M (henceforth
shortened to ``papers'') are directed graphs, so we make them
undirected by adding the reverse edge for every edge.  DGL's symmetric version of
arxiv and papers remove duplicates of edges with the same source and destination
while \systemname{} keeps them (i.e., \systemname{}'s graph has more edges
than DGL). products is already undirected. LDBC1000~\cite{ldbc-snb} is a synthetic graph with
generated features/labels used to show that \systemname{} can handle 
terabyte-scale graphs. We add random feature vectors of size 135 and make
80\% of 1\% of the vertices into training vertices. To partition the
graphs, we use a random edge-cut partition for both \systemname{} and DGL.
DGL supports METIS partitioning, but for a fair comparison with \systemname{},
we use random DGL for the main experiments.
As a special case, we show the runtime of DGL with METIS partitions (denoted
DGL-M) for papers on 32 hosts to discuss the differences with METIS.
Partitioning for \systemname{} occurs in a distributed manner when the graph is
loaded while DGL partitioning happens offline on a large memory machine
independent of the experimental machines for each rank count.

The Adam optimizer is used to update gradients. We use a 0.003 learning rate.
Dropout percentage for the GraphSAGE and GCN layers is 0.5.  Sampling is
with replacement for both training and testing.  We drop the last training
minibatch because that minibatch may contain a different number of vertices on
each rank since it is not guaranteed that every rank has the exact same number
of training vertices. PyTorch Distributed Data Parallel is used in both
\systemname{} and DistDGL to synchronize gradients.

\section{Additional Experimental Results and Analysis}

\begin{table*}[tbp!]
\footnotesize
\centering{}
\caption{Average runtime (seconds) among all ranks and breakdown of distributed
GNN training for 1 epoch for SAGE.  Prep time includes time to prepare
a subgraph: topology sampling, feature fetching, and export.}
\label{tbl:app_results1}
\begin{tabular}{l|l|l|rrrrrrr}
\toprule
                                    &                              &                                  & \multicolumn{7}{c}{\textbf{SAGE}}                                                                                                                                                                                                                               \\ \cline{4-10} 
                                    &                              &                                  & \multicolumn{4}{c|}{\textbf{Subgraph Prep}}                                                                                                        & \multicolumn{2}{c|}{\textbf{Compute}}                                & \multicolumn{1}{c}{\textbf{Epoch}}  \\
\textbf{Graph}                      & \textbf{Ranks}               & \textbf{System}                  & \multicolumn{1}{c}{\textbf{Topo}} & \multicolumn{1}{c}{\textbf{Feat}} & \multicolumn{1}{c|}{\textbf{Export}} & \multicolumn{1}{c|}{\textbf{Total}} & \multicolumn{1}{c}{\textbf{Fwd}} & \multicolumn{1}{c|}{\textbf{Bwd}} & \multicolumn{1}{c}{\textbf{Total}}  \\
\midrule
\multirow{9}{*}{\textbf{arxiv}}     & \multirow{3}{*}{\textbf{1}}  & \textbf{DGL}                     & 2.57                              & 0.23                              & \multicolumn{1}{r|}{-}               & \multicolumn{1}{r|}{2.80}           & 0.99                             & \multicolumn{1}{r|}{1.02}         & 4.96                                 \\
                                    &                              & \textbf{\systemnameabbr{}}       & 0.08                              & 0.13                              & \multicolumn{1}{r|}{0.13}            & \multicolumn{1}{r|}{0.39}           & 1.58                             & \multicolumn{1}{r|}{1.35}         & 3.45                                 \\
                                    &                              & \textbf{\systemnameabbrc{}}      & 0.07                              & 0.11                              & \multicolumn{1}{r|}{0.06}            & \multicolumn{1}{r|}{0.29}           & 1.70                             & \multicolumn{1}{r|}{1.43}         & 3.53                                 \\ \cline{2-10}
                                    & \multirow{3}{*}{\textbf{2}}  & \textbf{DGL}                     & 1.57                              & 0.66                              & \multicolumn{1}{r|}{-}               & \multicolumn{1}{r|}{2.23}           & 0.46                             & \multicolumn{1}{r|}{0.68}         & 3.45                                 \\
                                    &                              & \textbf{\systemnameabbr{}}       & 0.12                              & 0.10                              & \multicolumn{1}{r|}{0.07}            & \multicolumn{1}{r|}{0.30}           & 0.83                             & \multicolumn{1}{r|}{0.87}         & 2.06                                 \\
                                    &                              & \textbf{\systemnameabbrc{}}      & 0.09                              & 0.06                              & \multicolumn{1}{r|}{0.04}            & \multicolumn{1}{r|}{0.21}           & 0.59                             & \multicolumn{1}{r|}{0.75}         & 1.61                                 \\ \cline{2-10}
                                    & \multirow{3}{*}{\textbf{4}}  & \textbf{DGL}                     & 0.76                              & 0.33                              & \multicolumn{1}{r|}{-}               & \multicolumn{1}{r|}{1.10}           & 0.25                             & \multicolumn{1}{r|}{0.48}         & 1.87                                 \\
                                    &                              & \textbf{\systemnameabbr{}}       & 0.11                              & 0.10                              & \multicolumn{1}{r|}{0.04}            & \multicolumn{1}{r|}{0.26}           & 0.32                             & \multicolumn{1}{r|}{0.58}         & 1.19                                 \\
                                    &                              & \textbf{\systemnameabbrc{}}      & 0.11                              & 0.07                              & \multicolumn{1}{r|}{0.03}            & \multicolumn{1}{r|}{0.22}           & 0.34                             & \multicolumn{1}{r|}{0.44}         & 1.03                                 \\
\midrule
\multirow{12}{*}{\textbf{products}} & \multirow{3}{*}{\textbf{1}}  & \textbf{DGL}                     & 16.93                             & 2.29                              & \multicolumn{1}{r|}{-}               & \multicolumn{1}{r|}{19.22}          & 5.00                             & \multicolumn{1}{r|}{4.71}         & 29.22                                \\
                                    &                              & \textbf{\systemnameabbr{}}       & 0.64                              & 1.18                              & \multicolumn{1}{r|}{1.46}            & \multicolumn{1}{r|}{3.37}           & 5.52                             & \multicolumn{1}{r|}{4.88}         & 14.04                                \\
                                    &                              & \textbf{\systemnameabbrc{}}      & 0.41                              & 0.54                              & \multicolumn{1}{r|}{0.63}            & \multicolumn{1}{r|}{1.68}           & 4.71                             & \multicolumn{1}{r|}{4.07}         & 10.71                                \\ \cline{2-10}
                                    & \multirow{3}{*}{\textbf{2}}  & \textbf{DGL}                     & 11.74                             & 10.10                             & \multicolumn{1}{r|}{-}               & \multicolumn{1}{r|}{21.84}          & 2.90                             & \multicolumn{1}{r|}{4.04}         & 28.92                                \\
                                    &                              & \textbf{\systemnameabbr{}}       & 0.83                              & 1.00                              & \multicolumn{1}{r|}{0.91}            & \multicolumn{1}{r|}{2.79}           & 2.86                             & \multicolumn{1}{r|}{2.33}         & 8.11                                 \\
                                    &                              & \textbf{\systemnameabbrc{}}      & 0.71                              & 0.52                              & \multicolumn{1}{r|}{0.36}            & \multicolumn{1}{r|}{1.64}           & 2.38                             & \multicolumn{1}{r|}{2.34}         & 6.49                                 \\ \cline{2-10}
                                    & \multirow{3}{*}{\textbf{4}}  & \textbf{DGL}                     & 5.89                              & 4.24                              & \multicolumn{1}{r|}{-}               & \multicolumn{1}{r|}{10.13}          & 1.60                             & \multicolumn{1}{r|}{3.67}         & 15.47                                \\
                                    &                              & \textbf{\systemnameabbr{}}       & 0.52                              & 0.87                              & \multicolumn{1}{r|}{0.46}            & \multicolumn{1}{r|}{1.89}           & 1.41                             & \multicolumn{1}{r|}{1.54}         & 4.90                                 \\
                                    &                              & \textbf{\systemnameabbrc{}}      & 0.62                              & 0.50                              & \multicolumn{1}{r|}{0.22}            & \multicolumn{1}{r|}{1.36}           & 1.19                             & \multicolumn{1}{r|}{1.25}         & 3.87                                 \\ \cline{2-10}
                                    & \multirow{3}{*}{\textbf{8}}  & \textbf{DGL}                     & 3.21                              & 4.48                              & \multicolumn{1}{r|}{-}               & \multicolumn{1}{r|}{7.70}           & 0.84                             & \multicolumn{1}{r|}{3.30}         & 11.88                                \\
                                    &                              & \textbf{\systemnameabbr{}}       & 0.45                              & 0.73                              & \multicolumn{1}{r|}{0.31}            & \multicolumn{1}{r|}{1.52}           & 0.71                             & \multicolumn{1}{r|}{0.86}         & 3.12                                 \\
                                    &                              & \textbf{\systemnameabbrc{}}      & 0.65                              & 0.44                              & \multicolumn{1}{r|}{0.14}            & \multicolumn{1}{r|}{1.25}           & 0.61                             & \multicolumn{1}{r|}{0.86}         & 2.74                                 \\
\midrule
\multirow{16}{*}{\textbf{papers}}   & \multirow{3}{*}{\textbf{2}}  & \textbf{DGL}                     & 73.00                             & 77.41                             & \multicolumn{1}{r|}{-}               & \multicolumn{1}{r|}{150.41}         & 17.71                            & \multicolumn{1}{r|}{24.51}        & 193.96                               \\
                                    &                              & \textbf{\systemnameabbr{}}       & 3.98                              & 18.74                             & \multicolumn{1}{r|}{6.10}            & \multicolumn{1}{r|}{29.13}          & 16.27                            & \multicolumn{1}{r|}{17.51}        & 63.86                                \\
                                    &                              & \textbf{\systemnameabbrc{}}      & 5.33                              & 4.41                              & \multicolumn{1}{r|}{2.28}            & \multicolumn{1}{r|}{12.33}          & 13.28                            & \multicolumn{1}{r|}{15.51}        & 42.16                                \\ \cline{2-10}
                                    & \multirow{3}{*}{\textbf{4}}  & \textbf{DGL}                     & 34.27                             & 31.06                             & \multicolumn{1}{r|}{-}               & \multicolumn{1}{r|}{65.32}          & 8.74                             & \multicolumn{1}{r|}{18.49}        & 93.09                                \\
                                    &                              & \textbf{\systemnameabbr{}}       & 2.35                              & 7.20                              & \multicolumn{1}{r|}{3.78}            & \multicolumn{1}{r|}{13.64}          & 8.69                             & \multicolumn{1}{r|}{10.75}        & 33.61                                \\
                                    &                              & \textbf{\systemnameabbrc{}}      & 4.06                              & 2.73                              & \multicolumn{1}{r|}{1.19}            & \multicolumn{1}{r|}{8.14}           & 6.49                             & \multicolumn{1}{r|}{8.46}         & 23.62                                \\ \cline{2-10}
                                    & \multirow{3}{*}{\textbf{8}}  & \textbf{DGL}                     & 19.24                             & 38.00                             & \multicolumn{1}{r|}{-}               & \multicolumn{1}{r|}{57.24}          & 4.52                             & \multicolumn{1}{r|}{39.85}        & 101.91                               \\
                                    &                              & \textbf{\systemnameabbr{}}       & 1.37                              & 5.64                              & \multicolumn{1}{r|}{1.61}            & \multicolumn{1}{r|}{8.73}           & 3.96                             & \multicolumn{1}{r|}{6.77}         & 19.70                                \\
                                    &                              & \textbf{\systemnameabbrc{}}      & 3.28                              & 2.26                              & \multicolumn{1}{r|}{0.56}            & \multicolumn{1}{r|}{6.18}           & 3.21                             & \multicolumn{1}{r|}{4.14}         & 13.77                                \\ \cline{2-10}
                                    & \multirow{3}{*}{\textbf{16}} & \textbf{DGL}                     & 10.01                             & 26.57                             & \multicolumn{1}{r|}{-}               & \multicolumn{1}{r|}{36.58}          & 2.17                             & \multicolumn{1}{r|}{28.13}        & 67.03                                \\
                                    &                              & \textbf{\systemnameabbr{}}       & 0.76                              & 4.38                              & \multicolumn{1}{r|}{0.88}            & \multicolumn{1}{r|}{6.09}           & 1.96                             & \multicolumn{1}{r|}{3.34}         & 11.50                                \\
                                    &                              & \textbf{\systemnameabbrc{}}      & 2.02                              & 1.48                              & \multicolumn{1}{r|}{0.31}            & \multicolumn{1}{r|}{3.86}           & 1.71                             & \multicolumn{1}{r|}{2.69}         & 8.39                                 \\ \cline{2-10}
                                    & \multirow{4}{*}{\textbf{32}} & \textbf{DGL}                     & 5.64                              & 21.49                             & \multicolumn{1}{r|}{-}               & \multicolumn{1}{r|}{27.13}          & 1.08                             & \multicolumn{1}{r|}{13.66}        & 41.96                                \\
                                    &                              & \textbf{DGL-M}                   & 4.51                              & 12.55                             & \multicolumn{1}{r|}{-}               & \multicolumn{1}{r|}{17.06}          & 0.94                             & \multicolumn{1}{r|}{16.93}        & 35.02                                 \\
                                    &                              & \textbf{\systemnameabbr{}}       & 0.55                              & 2.76                              & \multicolumn{1}{r|}{0.45}            & \multicolumn{1}{r|}{3.79}           & 0.99                             & \multicolumn{1}{r|}{1.81}         & 6.64                                 \\
                                    &                              & \textbf{\systemnameabbrc{}}      & 1.41                              & 1.06                              & \multicolumn{1}{r|}{0.15}            & \multicolumn{1}{r|}{2.65}           & 0.86                             & \multicolumn{1}{r|}{2.17}         & 5.73                                 \\
\bottomrule
\end{tabular}
\end{table*}

\begin{table*}[tbp!]
\footnotesize
\centering{}
\caption{Average runtime (seconds) among all ranks and breakdown of distributed
GNN training for 1 epoch for GCN and GIN. Prep time for these is similar to
SAGE since the GNN layer differences will not change how sampling occurs. It
can be derived by subtracting the forward and backward times from the epoch
total.}
\label{tbl:app_results2}
\begin{tabular}{l|l|l|rrr|rrr}
\toprule
                                    &                              &                                  & \multicolumn{3}{c|}{\textbf{GCN}}                                                                          & \multicolumn{3}{c}{\textbf{GIN}}                                                                          \\ \cline{4-9} 
                                    &                              &                                  & \multicolumn{2}{c|}{\textbf{Compute}}                                & \multicolumn{1}{c|}{\textbf{Epoch}} & \multicolumn{2}{c|}{\textbf{Compute}}                                & \multicolumn{1}{c}{\textbf{Epoch}} \\
\textbf{Graph}                      & \textbf{Ranks}               & \textbf{System}                  & \multicolumn{1}{c}{\textbf{Fwd}} & \multicolumn{1}{c|}{\textbf{Bwd}} & \multicolumn{1}{c|}{\textbf{Total}} & \multicolumn{1}{c}{\textbf{Fwd}} & \multicolumn{1}{c|}{\textbf{Bwd}} & \multicolumn{1}{c}{\textbf{Total}} \\
\midrule
\multirow{9}{*}{\textbf{arxiv}}     & \multirow{3}{*}{\textbf{1}}  & \textbf{DGL}                     & 1.16                             & \multicolumn{1}{r|}{0.73}         & 4.95                                & 1.11                             & \multicolumn{1}{r|}{1.42}         & 5.78                               \\
                                    &                              & \textbf{\systemnameabbr{}}       & 0.84                             & \multicolumn{1}{r|}{0.75}         & 2.02                                & 1.20                             & \multicolumn{1}{r|}{1.42}         & 3.12                               \\
                                    &                              & \textbf{\systemnameabbrc{}}      & 1.05                             & \multicolumn{1}{r|}{0.93}         & 2.28                                & 1.38                             & \multicolumn{1}{r|}{1.62}         & 3.41                               \\ \cline{2-9}
                                    & \multirow{3}{*}{\textbf{2}}  & \textbf{DGL}                     & 0.52                             & \multicolumn{1}{r|}{0.61}         & 3.35                                & 0.76                             & \multicolumn{1}{r|}{0.80}         & 4.27                               \\
                                    &                              & \textbf{\systemnameabbr{}}       & 0.52                             & \multicolumn{1}{r|}{0.57}         & 1.43                                & 0.68                             & \multicolumn{1}{r|}{0.90}         & 1.99                               \\
                                    &                              & \textbf{\systemnameabbrc{}}      & 0.56                             & \multicolumn{1}{r|}{0.75}         & 1.58                                & 0.83                             & \multicolumn{1}{r|}{1.05}         & 2.26                               \\ \cline{2-9}
                                    & \multirow{3}{*}{\textbf{4}}  & \textbf{DGL}                     & 0.30                             & \multicolumn{1}{r|}{0.46}         & 1.93                                & 0.40                             & \multicolumn{1}{r|}{0.51}         & 2.06                               \\
                                    &                              & \textbf{\systemnameabbr{}}       & 0.31                             & \multicolumn{1}{r|}{0.47}         & 1.04                                & 0.35                             & \multicolumn{1}{r|}{0.51}         & 1.13                               \\
                                    &                              & \textbf{\systemnameabbrc{}}      & 0.29                             & \multicolumn{1}{r|}{0.38}         & 0.90                                & 0.36                             & \multicolumn{1}{r|}{0.53}         & 1.15                               \\
\midrule
\multirow{12}{*}{\textbf{products}} & \multirow{3}{*}{\textbf{1}}  & \textbf{DGL}                     & 7.76                             & \multicolumn{1}{r|}{4.24}         & 31.57                               & 5.15                             & \multicolumn{1}{r|}{7.24}         & 31.80                              \\
                                    &                              & \textbf{\systemnameabbr{}}       & 5.00                             & \multicolumn{1}{r|}{3.25}         & 12.11                               & 7.08                             & \multicolumn{1}{r|}{8.31}         & 19.05                              \\
                                    &                              & \textbf{\systemnameabbrc{}}      & 4.66                             & \multicolumn{1}{r|}{4.39}         & 11.00                               & 6.77                             & \multicolumn{1}{r|}{8.68}         & 17.51                              \\ \cline{2-9}
                                    & \multirow{3}{*}{\textbf{2}}  & \textbf{DGL}                     & 3.74                             & \multicolumn{1}{r|}{2.72}         & 27.23                               & 3.48                             & \multicolumn{1}{r|}{4.27}         & 28.67                              \\
                                    &                              & \textbf{\systemnameabbr{}}       & 2.68                             & \multicolumn{1}{r|}{2.29}         & 7.77                                & 3.67                             & \multicolumn{1}{r|}{4.68}         & 11.31                              \\
                                    &                              & \textbf{\systemnameabbrc{}}      & 2.27                             & \multicolumn{1}{r|}{1.96}         & 5.93                                & 3.32                             & \multicolumn{1}{r|}{4.51}         & 9.67                               \\ \cline{2-9}
                                    & \multirow{3}{*}{\textbf{4}}  & \textbf{DGL}                     & 2.05                             & \multicolumn{1}{r|}{2.87}         & 15.84                               & 2.54                             & \multicolumn{1}{r|}{2.36}         & 15.37                              \\
                                    &                              & \textbf{\systemnameabbr{}}       & 1.32                             & \multicolumn{1}{r|}{1.46}         & 4.72                                & 1.87                             & \multicolumn{1}{r|}{2.51}         & 6.59                               \\
                                    &                              & \textbf{\systemnameabbrc{}}      & 1.05                             & \multicolumn{1}{r|}{1.04}         & 3.54                                & 1.72                             & \multicolumn{1}{r|}{2.45}         & 5.63                               \\ \cline{2-9}
                                    & \multirow{3}{*}{\textbf{8}}  & \textbf{DGL}                     & 1.05                             & \multicolumn{1}{r|}{6.30}         & 16.07                               & 5.31                             & \multicolumn{1}{r|}{1.51}         & 15.15                              \\
                                    &                              & \textbf{\systemnameabbr{}}       & 0.64                             & \multicolumn{1}{r|}{0.73}         & 2.84                                & 0.96                             & \multicolumn{1}{r|}{1.44}         & 3.86                               \\
                                    &                              & \textbf{\systemnameabbrc{}}      & 0.55                             & \multicolumn{1}{r|}{0.74}         & 2.54                                & 0.88                             & \multicolumn{1}{r|}{1.33}         & 3.40                               \\
\midrule
\multirow{16}{*}{\textbf{papers}}   & \multirow{3}{*}{\textbf{2}}  & \textbf{DGL}                     & 22.29                            & \multicolumn{1}{r|}{21.63}        & 190.62                              & 21.54                            & \multicolumn{1}{r|}{23.86}        & 188.11                             \\
                                    &                              & \textbf{\systemnameabbr{}}       & 17.29                            & \multicolumn{1}{r|}{18.53}        & 64.67                               & 23.76                            & \multicolumn{1}{r|}{32.78}        & 89.22                              \\
                                    &                              & \textbf{\systemnameabbrc{}}      & 11.96                            & \multicolumn{1}{r|}{13.80}        & 38.85                               & 17.65                            & \multicolumn{1}{r|}{23.96}        & 55.21                              \\ \cline{2-9}
                                    & \multirow{3}{*}{\textbf{4}}  & \textbf{DGL}                     & 11.07                            & \multicolumn{1}{r|}{15.73}        & 93.61                               & 18.17                            & \multicolumn{1}{r|}{13.33}        & 99.93                              \\
                                    &                              & \textbf{\systemnameabbr{}}       & 7.93                             & \multicolumn{1}{r|}{11.31}        & 32.89                               & 11.40                            & \multicolumn{1}{r|}{20.29}        & 45.51                              \\
                                    &                              & \textbf{\systemnameabbrc{}}      & 5.71                             & \multicolumn{1}{r|}{7.30}         & 21.58                               & 9.43                             & \multicolumn{1}{r|}{13.59}        & 31.68                              \\ \cline{2-9}
                                    & \multirow{3}{*}{\textbf{8}}  & \textbf{DGL}                     & 5.65                             & \multicolumn{1}{r|}{39.09}        & 100.61                              & 37.85                            & \multicolumn{1}{r|}{7.28}         & 100.68                             \\
                                    &                              & \textbf{\systemnameabbr{}}       & 3.46                             & \multicolumn{1}{r|}{5.83}         & 18.00                               & 4.77                             & \multicolumn{1}{r|}{7.38}         & 20.97                              \\
                                    &                              & \textbf{\systemnameabbrc{}}      & 2.86                             & \multicolumn{1}{r|}{3.84}         & 13.06                               & 4.59                             & \multicolumn{1}{r|}{7.97}         & 18.82                              \\ \cline{2-9}
                                    & \multirow{3}{*}{\textbf{16}} & \textbf{DGL}                     & 2.84                             & \multicolumn{1}{r|}{29.73}        & 71.52                               & 29.44                            & \multicolumn{1}{r|}{4.27}         & 71.89                              \\
                                    &                              & \textbf{\systemnameabbr{}}       & 1.77                             & \multicolumn{1}{r|}{2.93}         & 11.09                               & 2.58                             & \multicolumn{1}{r|}{3.82}         & 12.54                              \\
                                    &                              & \textbf{\systemnameabbrc{}}      & 1.48                             & \multicolumn{1}{r|}{2.74}         & 7.98                                & 2.31                             & \multicolumn{1}{r|}{3.72}         & 10.17                              \\ \cline{2-9}
                                    & \multirow{4}{*}{\textbf{32}} & \textbf{DGL}                     & 1.43                             & \multicolumn{1}{r|}{14.87}        & 42.96                               & 13.67                            & \multicolumn{1}{r|}{2.29}         & 43.33                              \\
                                    &                              & \textbf{DGL-M}                   & 1.18                             & \multicolumn{1}{r|}{17.08}        & 33.88                               & 16.88                            & \multicolumn{1}{r|}{2.13}         & 35.52                               \\
                                    &                              & \textbf{\systemnameabbr{}}       & 0.89                             & \multicolumn{1}{r|}{2.20}         & 7.00                                & 1.35                             & \multicolumn{1}{r|}{2.34}         & 7.57                               \\
                                    &                              & \textbf{\systemnameabbrc{}}      & 0.76                             & \multicolumn{1}{r|}{1.94}         & 5.36                                & 1.18                             & \multicolumn{1}{r|}{1.95}         & 5.77                               \\
\bottomrule
\end{tabular}
\end{table*}

\subsection{Full Set of Results}
\label{app:full_results}

\Cref{tbl:app_results1} and \Cref{tbl:app_results2} show the full set of scaling
results for the SAGE, GCN, and GIN experiments in the main paper. Specifically,
it includes scaling for up to 32 ranks and the forward/backward breakdown for
GCN and GIN both of which aren't in the main paper due to space constraints. 1
rank papers100M is not included due to memory constraints. Speedup numbers in
the main paper use the entirety of these numbers for their computation and not
just the ones in \Cref{tbl:results}.

\Cref{tbl:cache_init} shows the aggregation cache initialization time. This
is a one time cost that can be amortized over multiple epochs, and the cost
decreases as the number of ranks increases due to additional compute power.

\begin{table}[h!]
\small
\centering
\caption{Aggregation cache initialization time (in seconds) for various graphs
for the SAGE run.}
\label{tbl:cache_init}
\begin{tabular}{r|rrrrrr}
\toprule
                  & \multicolumn{6}{c}{{\textbf{Ranks}}}                                                          \\ \cline{2-7} 
\textbf{}         & \textbf{1}                     & \textbf{2} & \textbf{4} & \textbf{8} & \textbf{16} & \textbf{32} \\
\midrule
\textbf{arxiv}    & 0.44                           & 0.24       & 0.15       & -          & -           & -           \\
\textbf{products} & 22.65                          & 11.20      & 5.86       & 3.45       & -           & -           \\
\textbf{papers}   & -                              & 320.69     & 176.11     & 87.95      & 45.69       & 23.22       \\
\bottomrule
\end{tabular}
\end{table}

\subsection{Discussion on Load Imbalance}
\label{app:imbalance}

What follows is an unabridged discussion with detailed experimental results on
load imbalance that provides more detail than the main paper. Portions of this
text are found in the main paper.

\begin{table}[t]
\footnotesize
\centering{}
\caption{
Average subgraph preparation and forward phase time (P+F) and
the average difference (Rng.) in P+F between the fastest and slowest rank
\emph{for one minibatch} in an epoch for various graphs, rank (Rk.), and
model configurations.
}
\label{tbl:imbalance}
\begin{tabular}{llllrrrrr}
\toprule
                                                                   &                                                 &                                                        & \multicolumn{2}{c}{\textbf{SAGE}}                                              & \multicolumn{2}{c}{\textbf{GCN}}                     & \multicolumn{2}{c}{\textbf{GIN}}                    \\
\multicolumn{1}{l|}{\textbf{Graph}}                                & \multicolumn{1}{l|}{\textbf{Rk.}}               & \multicolumn{1}{l|}{\textbf{Sys.}}                     & \multicolumn{1}{l}{\textbf{P+F}} & \multicolumn{1}{l|}{\textbf{Rng.}} & \multicolumn{1}{l}{\textbf{P+F}} & \multicolumn{1}{l|}{\textbf{Rng.}} & \multicolumn{1}{l}{\textbf{P+F}} & \multicolumn{1}{l}{\textbf{Rng.}} \\
\midrule
\multicolumn{1}{l|}{\multirow{4}{*}{arxiv}}                        & \multicolumn{1}{l|}{\multirow{2}{*}{2}}         & \multicolumn{1}{l|}{DGL}                               & 0.06                    & \multicolumn{1}{r|}{0.01}  & 0.06                    & \multicolumn{1}{r|}{0.01}  & 0.08                    & 0.00                      \\
\multicolumn{1}{l|}{}                                              & \multicolumn{1}{l|}{}                           & \multicolumn{1}{l|}{\systemnameabbr{}}                 & 0.03                    & \multicolumn{1}{r|}{0.00}  & 0.02                    & \multicolumn{1}{r|}{0.00}  & 0.02                    & 0.00                      \\ \cline{2-9}
\multicolumn{1}{l|}{}                                              & \multicolumn{1}{l|}{\multirow{2}{*}{4}}         & \multicolumn{1}{l|}{DGL}                               & 0.06                    & \multicolumn{1}{r|}{0.01}  & 0.07                    & \multicolumn{1}{r|}{0.02}  & 0.07                    & 0.00                      \\
\multicolumn{1}{l|}{}                                              & \multicolumn{1}{l|}{}                           & \multicolumn{1}{l|}{\systemnameabbr{}}                 & 0.03                    & \multicolumn{1}{r|}{0.01}  & 0.02                    & \multicolumn{1}{r|}{0.01}  & 0.03                    & 0.00                      \\
\midrule
\multicolumn{1}{l|}{\multirow{6}{*}{product}}                      & \multicolumn{1}{l|}{\multirow{2}{*}{2}}         & \multicolumn{1}{l|}{DGL}                               & 0.26                    & \multicolumn{1}{r|}{0.02}  & 0.25                    & \multicolumn{1}{r|}{0.01}  & 0.25                    & 0.00                      \\
\multicolumn{1}{l|}{}                                              & \multicolumn{1}{l|}{}                           & \multicolumn{1}{l|}{\systemnameabbr{}}                 & 0.06                    & \multicolumn{1}{r|}{0.00}  & 0.06                    & \multicolumn{1}{r|}{0.01}  & 0.07                    & 0.00                      \\ \cline{2-9}
\multicolumn{1}{l|}{}                                              & \multicolumn{1}{l|}{\multirow{2}{*}{4}}         & \multicolumn{1}{l|}{DGL}                               & 0.24                    & \multicolumn{1}{r|}{0.07}  & 0.27                    & \multicolumn{1}{r|}{0.07}  & 0.27                    & 0.01                      \\
\multicolumn{1}{l|}{}                                              & \multicolumn{1}{l|}{}                           & \multicolumn{1}{l|}{\systemnameabbr{}}                 & 0.07                    & \multicolumn{1}{r|}{0.01}  & 0.07                    & \multicolumn{1}{r|}{0.01}  & 0.08                    & 0.01                      \\ \cline{2-9}
\multicolumn{1}{l|}{}                                              & \multicolumn{1}{l|}{\multirow{2}{*}{8}}         & \multicolumn{1}{l|}{DGL}                               & 0.36                    & \multicolumn{1}{r|}{0.22}  & 0.41                    & \multicolumn{1}{r|}{0.40}  & 0.57                    & 0.01                      \\
\multicolumn{1}{l|}{}                                              & \multicolumn{1}{l|}{}                           & \multicolumn{1}{l|}{\systemnameabbr{}}                 & 0.09                    & \multicolumn{1}{r|}{0.01}  & 0.09                    & \multicolumn{1}{r|}{0.01}  & 0.10                    & 0.01                      \\
\midrule
\multicolumn{1}{l|}{\multirow{11}{*}{paper}}                       & \multicolumn{1}{l|}{\multirow{2}{*}{2}}         & \multicolumn{1}{l|}{DGL}                               & 0.29                    & \multicolumn{1}{r|}{0.02}  & 0.29                    & \multicolumn{1}{r|}{0.03}  & 0.28                    & 0.00                      \\
\multicolumn{1}{l|}{}                                              & \multicolumn{1}{l|}{}                           & \multicolumn{1}{l|}{\systemnameabbr{}}                 & 0.08                    & \multicolumn{1}{r|}{0.01}  & 0.08                    & \multicolumn{1}{r|}{0.02}  & 0.09                    & 0.01                      \\ \cline{2-9}
\multicolumn{1}{l|}{}                                              & \multicolumn{1}{l|}{\multirow{2}{*}{4}}         & \multicolumn{1}{l|}{DGL}                               & 0.25                    & \multicolumn{1}{r|}{0.07}  & 0.26                    & \multicolumn{1}{r|}{0.07}  & 0.29                    & 0.00                      \\
\multicolumn{1}{l|}{}                                              & \multicolumn{1}{l|}{}                           & \multicolumn{1}{l|}{\systemnameabbr{}}                 & 0.08                    & \multicolumn{1}{r|}{0.02}  & 0.07                    & \multicolumn{1}{r|}{0.02}  & 0.08                    & 0.02                      \\ \cline{2-9}
\multicolumn{1}{l|}{}                                              & \multicolumn{1}{l|}{\multirow{2}{*}{8}}         & \multicolumn{1}{l|}{DGL}                               & 0.42                    & \multicolumn{1}{r|}{0.42}  & 0.42                    & \multicolumn{1}{r|}{0.41}  & 0.63                    & 0.01                      \\
\multicolumn{1}{l|}{}                                              & \multicolumn{1}{l|}{}                           & \multicolumn{1}{l|}{\systemnameabbr{}}                 & 0.09                    & \multicolumn{1}{r|}{0.02}  & 0.08                    & \multicolumn{1}{r|}{0.02}  & 0.09                    & 0.01                      \\ \cline{2-9}
\multicolumn{1}{l|}{}                                              & \multicolumn{1}{l|}{\multirow{2}{*}{16}}        & \multicolumn{1}{l|}{DGL}                               & 0.53                    & \multicolumn{1}{r|}{0.57}  & 0.57                    & \multicolumn{1}{r|}{0.63}  & 0.92                    & 0.01                      \\
\multicolumn{1}{l|}{}                                              & \multicolumn{1}{l|}{}                           & \multicolumn{1}{l|}{\systemnameabbr{}}                 & 0.11                    & \multicolumn{1}{r|}{0.03}  & 0.11                    & \multicolumn{1}{r|}{0.02}  & 0.12                    & 0.01                      \\ \cline{2-9}
\multicolumn{1}{l|}{}                                              & \multicolumn{1}{l|}{\multirow{3}{*}{32}}        & \multicolumn{1}{l|}{DGL}                               & 0.78                    & \multicolumn{1}{r|}{0.63}  & 0.78                    & \multicolumn{1}{r|}{0.66}  & 1.14                    & 0.01                      \\
\multicolumn{1}{l|}{}                                              & \multicolumn{1}{l|}{}                           & \multicolumn{1}{l|}{DGL-M}                             & 0.50                    & \multicolumn{1}{r|}{0.75}  & 0.46                    & \multicolumn{1}{r|}{0.73}  & 0.92                    & 0.01                      \\
\multicolumn{1}{l|}{}                                              & \multicolumn{1}{l|}{}                           & \multicolumn{1}{l|}{\systemnameabbr{}}                 & 0.13                    & \multicolumn{1}{r|}{0.03}  & 0.13                    & \multicolumn{1}{r|}{0.04}  & 0.14                    & 0.03                      \\
\bottomrule
\end{tabular}
\end{table}

Observe in \Cref{tbl:app_results1} and \Cref{tbl:app_results2} that (1) DGL's
backward phase time for SAGE and GCN and (2) DGL's forward phase time for GIN
are both significantly higher than \systemnameabbr{}'s for some configurations
(e.g., products and papers past 4 ranks): the reason for this is very heavy load
imbalance in execution.

\Cref{tbl:imbalance} illustrates load imbalance 
by showing the average difference between the fastest and slowest rank when it
comes to subgraph preparation and forward phase time.  This quantity is
important because it is a measure of how long the entire system may be bottlenecked
if there is a barrier during execution as all ranks would have to wait for the
slowest rank. The backward phase for all models has a logical barrier due to the
synchronization of weight gradients. Therefore, with high imbalance, the
backward phase time recorded in the experiments will grow due to wait time. 

We can explain the reason for DGL's performance with this information. For
products and papers past 4 ranks, the imbalance among ranks gets close or even
surpasses the average prep and forward pass time. This indicates heavy imbalance
that results in the high backward phase time shown in \Cref{tbl:app_results1}
and \Cref{tbl:app_results2} for SAGE and GCN. For GIN, the forward phase time is
instead very high for DGL compared to \systemnameabbr{}, and the imbalance shown
in \Cref{tbl:imbalance} is actually low. The reason for this is also imbalance:
Torch Distributed Data Parallel (DDP, the mechanism used to synchronize models
in our experiments) adds a barrier before the \emph{forward} phase in
GIN (Our testing suggests that DDP inserts this barrier if there are
many learnable parameters in the model. GIN has more parameters to learn than
SAGE and GCN.), so the imbalance present in the sampling and feature fetching
(the main source of imbalance in these runs) causes all ranks to wait for the
straggler host before forward rather than at backward.  The result is that the
imbalance for prep and forward phase is low due to the existence of the barrier,
but the imbalance overhead is now reflected in GIN's very high time in
\Cref{tbl:app_results2}. The reason that DGL-M does not result in a larger
performance improvement over DGL can be explained with imbalance as well.  METIS
partitions result in more relative imbalance than random partitions as shown in
\Cref{tbl:imbalance}, so the backward time for SAGE/GCN and forward time in GIN
in \Cref{tbl:app_results1} and \Cref{tbl:app_results2} for DGL-M is higher than
that of normal DGL.

\systemname{} does not have these issues: \Cref{tbl:imbalance} shows all 
configurations are relatively balanced.  
Any sampling imbalance only occurs when fetching macrobatches rather than for
every minibatch in the macrobatch.  Macrobatching \emph{mitigates} the effect of
imbalance because sampling does not occur past the first minibatch of a
macrobatch.  When minibatches are sampled one at a time like in DGL, sampling
imbalance reoccurs for every minibatch, \emph{amplifying} the imbalance.
Therefore, a major takeaway is that macrobatching improves runtime by mitigating
load imbalance overheads.

\section{Links to Used Resources}

The OGBN datasets can be found here:
\url{https://ogb.stanford.edu/docs/nodeprop/}

DGL code can be found here:
\url{https://github.com/dmlc/dgl}

LDBC can be found here:
\url{https://ldbcouncil.org/benchmarks/snb/}

The MKL package we used is 2022.1.0, hc2b9512\_224, and can be found here:
\url{https://anaconda.org/anaconda/mkl/files}

\end{document}